\documentclass[10pt,oneside,leqno, twocolumn]{article}

\usepackage[a4paper, top=1.5cm, left=1.5cm, bottom=1.5cm, right=1.5cm]{geometry}
\RequirePackage{amsthm,amsmath,amsfonts,amssymb}
\RequirePackage[authoryear, round]{natbib}%% uncomment this for author-year citations
\RequirePackage[colorlinks,citecolor=blue,urlcolor=blue,linkcolor=blue]{hyperref}%% uncomment this for coloring bibliography citations and linked URLs
\RequirePackage{graphicx}%% uncomment this for including figures

\usepackage[T1]{fontenc} %for < and >
\usepackage[utf8]{inputenc}
\usepackage[english]{babel}
\usepackage{bm}
\usepackage{enumitem}
\usepackage[dvipsnames]{xcolor}
\usepackage[section]{placeins}
\usepackage{mathtools}
\usepackage{tikz} %for plots from R tikzDevice

\usepackage{multirow} % for spanning multiple rows in table
\usepackage{caption}  % for putting two tables next to each other and still having good captions
\usepackage{subcaption} %plots next to each other

\usepackage{nameref}
\usepackage{easyReview}
\usepackage{marginnote}

\usepackage{authblk}
\usepackage{abstract}
\usepackage{orcidlink}

\newcommand{\Prob}{\mathbb{P}}

\renewenvironment{abstract}
 {\small
  \begin{center}
  \bfseries \abstractname\vspace{-.5em}\vspace{0pt}
  \end{center}
  \list{}{%
    \setlength{\leftmargin}{20mm}% <---------- CHANGE HERE
    \setlength{\rightmargin}{\leftmargin}%
  }%
  \item\relax}
 {\endlist}

\newenvironment{localsize}[1]
{%
  \clearpage
  \let\orignewcommand\newcommand
  \let\newcommand\renewcommand
  \makeatletter
  \input{size#1.clo}%
  \makeatother
  \let\newcommand\orignewcommand
}
{%
  \clearpage
}

\title{Sources of Uncertainty in Supervised Machine Learning -- A Statisticians' View}
\author{Cornelia Gruber\thanks{cornelia.gruber@stat.uni-muenchen.de}}
\author{Patrick Oliver Schenk}
\author{Malte Schierholz}
\author{Frauke Kreuter}
\author{G{\"{o}}ran Kauermann}
\affil{Department of Statistics, Ludwig-Maximilians-University Munich, Ludwigstr. 33,  80539 Munich, Germany}

\date{}

\begin{document}

\twocolumn[
 \begin{@twocolumnfalse}
    \maketitle
\begin{abstract}
Supervised machine learning and predictive models have achieved an impressive standard today, enabling us to answer questions that were inconceivable a few years ago. Besides these successes, it becomes clear, that beyond pure prediction, which is the primary strength of most supervised machine learning algorithms, the quantification of uncertainty is relevant and necessary as well. 
However, before quantification is possible, types and sources of uncertainty need to be defined precisely.
While first concepts and ideas in this direction have emerged in recent years, this paper adopts a conceptual, basic science perspective and examines possible sources of uncertainty. By adopting the viewpoint of a statistician, we discuss the concepts of aleatoric and epistemic uncertainty, which are more commonly associated with machine learning. The paper aims to formalize the two types of uncertainty and demonstrates that sources of uncertainty are miscellaneous and can not always be decomposed into aleatoric and epistemic. Drawing parallels between statistical concepts and uncertainty in machine learning, we emphasise the role of data and their influence on uncertainty.
    \newline
    \newline
    {\bf Keywords:}
uncertainty; supervised machine learning; aleatoric uncertainty; epistemic uncertainty; label noise; measurement error; omitted variables; non i.i.d. data; distribution shift; missing data; data-centric machine learning.
    \newline
    \newline

\end{abstract}
  \end{@twocolumnfalse}
]
\saythanks

%\tableofcontents
%\newpage

%%%% Main text entry area:
% Kapitel 1
\section{Introduction}
In recent years, the predictions of machine learning algorithms, especially in the field of supervised learning, have become increasingly successful and can now achieve an impressive degree of accuracy. Especially in areas where not only accuracy but also reliability of predictions is crucial, the analysis and quantification of uncertainties is moving more into focus. For example, a physician would want to know whether a given patient is likely to have a critical condition, say sepsis, and would not just want to know the overall predictive accuracy of a classifier. 
Furthermore, an uncertainty-aware model could abstain from predictions if a certain level of uncertainty is exceeded \citep{kompa2021SecondOpinionNeeded}.
In addition to medical diagnosis and treatment \citep{ching2018OpportunitiesObstaclesDeep, defauw2018ClinicallyApplicableDeep, kompa2021SecondOpinionNeeded, leibig2017LeveragingUncertaintyInformation, qayyum2021SecureRobustMachine, valen2022QuantifyingUncertaintyMachine}, the application of uncertainty-aware models extends to many more fields where machine learning is applied today, including weather forecasting \citep{scher2018PredictingWeatherForecast}, engineering or industrial processes \citep{pan2021IntegratedDataKnowledge},  autonomous driving \citep{schwarting2018PlanningDecisionMakingAutonomous} or remote sensing based on satellite images \citep{zhu2017DeepLearningRemote}.

% More formally, yet oversimplifying a diverse literature \citep[e.g.,][]{abdar2021ReviewUncertaintyQuantification, hullermeier2021AleatoricEpistemicUncertainty}, machine learning methods that purport to quantify uncertainty aim to output a probability distribution $P(Y | x)$ over all possible outcomes of $y$ (e.g., sepsis: yes/no) or related metrics that could be derived from this distribution. $x$ is an input (vector) of features that might be observed in the future.
%an individual who might get observed in the future. If $y$ is binary, the problem simplifies: knowing the probability $\pi = P(Y = 1 | x)$ is equivalent to knowing the full distribution.

More formally, we consider the generic prediction of $Y | X$ where $X$ is an input (vector) of features that might be observed in the future and $Y$ is the true outcome of interest. This framework includes settings from predicting a simple coin toss ($X = \emptyset $ and $Y = \{0,1\}$), image classification ($X = \mathbb{R}^{H \times W \times 3}$ and, e.g., $Y = \{cat, dog\}$), to natural language processing ($X = V^l$ and Y = $\{positive, neutral, negative\}$ where $X$ is a text with $l$ tokens out of vocabulary $V$ and $Y$ is the sentiment of the text). We note that no assumption on the prediction model itself is made.
%More formally, let $X$ be an input (vector) of features that might be observed in the future and let $Y$ be the true outcome of interest (e.g., sepsis: yes/no).
Loosely speaking and oversimplifying a rich and diverse literature on uncertainty quantification \citep[e.g.,][]{abdar2021ReviewUncertaintyQuantification, hullermeier2021AleatoricEpistemicUncertainty}, machine learning methods that quantify (predictive) uncertainty aim to output a probability distribution $P(Y | x)$, or related metrics that could be derived from this distribution. Ideally, the relation between $X$ and $Y$ would be deterministic, such that for a future unit with input value $x$ the true value $y$ can be predicted with certainty, $P(Y = y | x) = 1$, ensuring that appropriate decisions can be made. Whenever this relation is not deterministic -- i.e., in many applied situations -- practitioners are left with uncertainty about the true value of $y$. Recent literature \citep{senge2014ReliableClassificationLearning, hullermeier2021AleatoricEpistemicUncertainty} distinguishes between two types of uncertainty: \textit{Epistemic uncertainty} entertains the idea that one could reduce the uncertainty surrounding $P(Y | x)$, for example by adding additional input features, more training observations, or with better knowledge about the true model, thus allowing for better and more reliable decision-making if a decision can be postponed. \textit{Aleatoric uncertainty}, conversely, refers to unavoidable, irreducible uncertainty, due to randomness inherent in a system.

While the usefulness of considering the different types of uncertainty for supervised machine learning may be apparent, the development of an underlying theory has not yet come to a commonly accepted standard.
Often, neither the definition of uncertainty nor the possible causes of uncertainty in machine learning applications are consistent and clearly stated. In fact, a large body of literature on uncertainty in machine learning has focused on various methods to quantify aleatoric and epistemic uncertainty, while little attention has been paid to the underlying sources of uncertainty, nor to the impact of data quality on the uncertainty of a prediction. 

Against this background, we believe that precise definitions and detailed transdisciplinary discussions about objectives and challenges of the field will be necessary to make scientific progress. Statistics as the discipline of reasoning with data, while facing randomness, has much to offer in this regard. Our contribution towards a more unified theory is threefold: First, we discuss different types of uncertainty, point to diverse statistical literatures that can inform the debate, and provide, to the extent possible, mathematically sound definitions from a statistical perspective.

Second, we call into question two main assumptions that are widespread in the literature on uncertainty quantification. We argue that a ``simple'' decomposition into aleatoric and epistemic uncertainty -- as various methods for uncertainty quantification do \citep[e.g.,][]{gal2016UncertaintyDeepLearning, depeweg2018DecompositionUncertaintyBayesian} -- does not do justice to the overarching problem of multiple sources of uncertainty \citep[see also][]{bengs2022pitfalls, wimmer2023QuantifyingAleatoricEpistemic, schweighofer2023IntroducingImprovedInformationTheoretic}. In addition, we refute a common perspective 
(e.g., \citeauthor{kendall2017WhatUncertaintiesWe}, \citeyear{kendall2017WhatUncertaintiesWe}, pp.~1-2; \citeauthor{hullermeier2021AleatoricEpistemicUncertainty}, \citeyear{hullermeier2021AleatoricEpistemicUncertainty}, p.~465),
 which states that model uncertainty, a part of epistemic uncertainty, can safely be ignored if learners that have universal approximation capabilities are being used, like nearest neighbour classification and (deep) neural networks. As we will show, that perspective rests on the idea that all the relevant features are always available and without measurement errors; in many applications this requirement is not fulfilled. 

This, finally, leads us to emphasize data production processes, missing data, and data quality as distinct sources of uncertainty, topics that have been largely ignored in the uncertainty quantification literature. If the data production process changes during data collection or between model training and its deployment, one needs to be extremely careful about making well-calibrated predictions. Only if a given dataset, and the model trained from it, is a good-enough virtual representation of the real world, can one hope to use it to make meaningful decisions about the real world. We, therefore, put \textit{data} as a possible source of uncertainty in the foreground.

The paper is structured as follows.
Section~\ref{sec:related-work} reviews the literature on uncertainty quantification as well as aleatoric and epistemic uncertainties. We illustrate the widespread usage and the relevance of these concepts by mentioning some practical applications.  
In Section~\ref{sec:sources-uncertainty} we provide statistical definitions of \textit{aleatoric uncertainty} and \textit{epistemic uncertainty}, including \textit{estimation uncertainty}, and \textit{model uncertainty}. We also draw parallels between uncertainty and classical statistics, such as the bias-variance decomposition or statistical reasoning about misspecified models. 
Section~\ref{sec:role.of.data} extends the statistical approach towards including latent quantities, which allows us to discuss uncertainties due to omitted variables or error-prone measurements. 
In Section~\ref{sec:data-uncertainty}, we take the data production process into account and relate uncertainties to the total survey error, a well-established framework tracing the different sources of uncertainty. Furthermore, uncertainties occurring due to unknown changes in the real world after a model has been trained, the deployment phase, are looked at.  
Section~\ref{sec:conclusion} concludes the paper.

% Kapitel 2
\section{Related Work and Applications}
\label{sec:related-work}
\subsection{Aleatoric and Epistemic Uncertainty in Supervised Machine Learning}

\citet{hullermeier2021AleatoricEpistemicUncertainty} provide an exhaustive survey of algorithms beyond classical supervised machine learning that handle aleatoric and epistemic uncertainty. The authors assume a correctly specified hypothesis space and make assumptions about the data generating process. 
According to the authors, in addition to aleatoric uncertainty, there is uncertainty in the choice of the model type (model uncertainty) and uncertainty in how well the fitted model can approximate the optimal model (approximation uncertainty). 
They, however, do not cover sources of uncertainty that occur while collecting data or when deploying the model to new data.
Not accounting for possible distribution shifts or discrepancies between training and test data can lead to even greater uncertainty and needs to be considered when applying or deploying predictive models in practice. 

\citet{abdar2021ReviewUncertaintyQuantification} extensively review uncertainty quantification methods for deep learning, mainly focusing on Bayesian approximations and ensemble techniques, and list advantages and disadvantages of the most commonly used uncertainty quantification methods. Additionally, a list of papers applying uncertainty quantification methods is presented.

\citet{gawlikowski2023SurveyUncertaintyDeep} extend previous definitions of aleatoric and epistemic uncertainty and propose aleatoric uncertainty to arise from errors and noise in measurement, and epistemic uncertainty to arise from variability in real-world situations, unknown data, errors during training, or errors in the model structure. Later, they present methods for quantifying aleatoric or epistemic uncertainty in deep learning models. However, the presented methods do not distinguish between the different sources of uncertainty.

We also refer to \citet{psaros2023UncertaintyQuantificationScientific, valdenegro-toro2022DeeperLookAleatoric, zhou2022SurveyEpistemicModel} for further aspects of aleatoric and epistemic uncertainty in machine learning. 

A quite general perspective on uncertainty quantification for deep learning in medical decision-making is provided by \citet{begoli2019NeedUncertaintyQuantification}. They suggest that to quantify a model's uncertainty, uncertainty in all steps of the decision model must be considered.  The components that play a role in the uncertainty of such a model are the data quality, i.e., how well the collected data represent real-world phenomena, the quality of the labelling process, the suitability of the model itself, as well as the performance on operational data. 
\citet{begoli2019NeedUncertaintyQuantification} also point to the need for a principled and formal discipline of uncertainty quantification, which we will address below.

We add to the literature by being more formal and pointing to shortcomings in existing definitions of aleatoric and epistemic uncertainty.
\subsection{Why Considering Uncertainties Helps}
\label{subsec:why.considering.uncertainties.helps}

Uncertainty quantification is particularly useful if humans are part of the decision-making process, who could use the uncertainty estimates to better assess the extent to which certain predictions can be trusted. 
In classification with a reject option, for example, the algorithm refrains from a prediction when an observation is hard to classify, e.g., because some confidence level is below a threshold, and instead reports a warning \citep{bartlett2008ClassificationRejectOption}. This method is also known as \textit{selective classification} \citep{el-yaniv2010FoundationsNoisefreeSelective}. 

Quantifying aleatoric and epistemic uncertainty separately showed to add value compared to a joint treatment in many specific tasks. For instance, knowledge of epistemic uncertainty is useful in active learning.
In situations where a huge amount of unlabelled data exists, but labelling is expensive, active learning is a promising strategy to build an accurate and cost-efficient model iteratively. The instances to be labelled next should be the ones that improve the model the most after retraining \citep{aggarwal2014ActiveLearningSurvey}.
\citet{nguyen2022HowMeasureUncertainty} show that labelling instances with high epistemic uncertainty is superior to random sampling as well as sampling based on entropy.
Thus, distinguishing types of uncertainty leads to a greater improvement of the algorithm than considering a general uncertainty measure.  
As expected, adding instances with high aleatoric uncertainty to the pool of labelled data showed not to be helpful.
This aligns with the understanding that epistemic uncertainty relates to a lack of knowledge, which can be reduced with informative samples.
The advantages of reducing epistemic uncertainty for active learning are also presented in \citet{gal2017DeepBayesianActive}.

% \rtwo{On the other hand, the paper touches upon too many issues that are very superficially mentioned in passing and do not fit into the paper. For example, the brief paragraph on adversarial learning on page 3 Is distracting and contributes very little to the main theme of the paper.
% }
% Another promising application for uncertainty-aware models is detecting adversarial examples, i.e., crafted samples where the model gives a highly confident, but incorrect, prediction. 
% Often those samples are designed in a way to be unremarkable to the human eye while still misleading the model by, e.g., adding specific noise to the example. Models that do not integrate uncertainty would simply output high predicted scores for one particular, but wrong, label.
% \citet{smith2018UnderstandingMeasuresUncertainty} examined models
% that quantify uncertainty and showed that they are able to identify such adversarial inputs and are therefore more robust and harder to fool. 

As suggested by \citet{kendall2017WhatUncertaintiesWe}, model robustness to noisy data can also be achieved by explicitly formulating uncertainty in the loss function. This ensures that inputs for which the model has learned to predict a high degree of uncertainty will have a lower impact on the loss.

Additionally, there were multiple successful applications of out-of-distribution (OOD) detection with models that distinguish aleatoric and epistemic uncertainty \citep{malinin2018PredictiveUncertaintyEstimation, mukhoti2023DeepDeterministicUncertainty, ren2022OutofDistributionDetectionSelective}.
As predictive models often deteriorate when confronted with OOD inputs, detecting such improper inputs is of high interest. 
Ideally, a model should return high uncertainty estimates for such inputs and be able to distinguish between uncertainty due to noise or class overlap, i.e., aleatoric uncertainty, and uncertainty due to unknown inputs \citep{malinin2018PredictiveUncertaintyEstimation}.

Further applications of uncertainty quantification in machine learning include adversarial learning \citep{smith2018UnderstandingMeasuresUncertainty}, image segmentation and depth regression \cite{kendall2017WhatUncertaintiesWe}.

The usefulness of uncertainty quantification in those applications is evident. However, a comprehensive understanding of types and sources of uncertainty needs to be ensured for a trustworthy application.

The rest of the paper considers the problem of uncertainty through a statistical lens and aims to broaden the view by formalizing the distinction between aleatoric and epistemic uncertainty and listing further potential sources where uncertainty may occur in the context of machine learning.

% \gk{Indeed, if we would know how much aleatoric uncertainty there is, this would help a lot since it keeps us from overfitting. 
% There is a nice paper. 
% They show that with nearest neighbours you can indeed estimate the aleatoric variance. But that's an asympotic result and my simulations in this direction indicated, that asymptotic here really means asymptotic. Still, we should mention this paper.
% \url{https://doi.org/10.1214/18-EJS1438}
% }

%Kapitel 3
\section{Sources of Uncertainty}
\label{sec:sources-uncertainty}

\subsection{Aleatoric and Epistemic Uncertainty}
\label{sec:aleatoric_epistemic_uncertainty}

As the terms ``aleatory''/``aleatoric'' and ``epistemology''/``epistemic'' have no precise definition (i.e., what is knowledge, what is reducible), we will first look at their etymology and historical development. Then we will provide a clear definition in statistical terms, which allows us to point to the shortcomings of existing definitions. 
Those terms were already used to describe different historical perspectives on the concept of probability, spanning the time between Blaise Pascal (*1623, \textdagger1654) and Pierre-Simon Laplace (*1749, \textdagger1827). These terms then developed into the nowadays common terms ``aleatoric'' and ``epistemic''. An aleatoric notion considers probability as the stable frequency of an event, while the epistemic notion of probability relates to the degree of belief \citep{hacking1975EmergenceProbabilityPhilosophical}. 
This differentiation can be understood as an early debate between frequentist and Bayesian views, respectively.
Later on, the terms were used to describe uncertainty in modelling and it was proposed that the total uncertainty decomposes\phantomsection\label{decomposes} into \textit{aleatoric} and \textit{epistemic} uncertainty \citep{HORA1996217, kiureghian2009AleatoryEpistemicDoes}. 
Aleatoric uncertainty is thereby defined as uncertainty arising from the inherent randomness of an event, while epistemic uncertainty expresses uncertainty due to a lack of knowledge.

Those terms then found their way into machine learning terminology \citep{senge2014ReliableClassificationLearning, gal2016UncertaintyDeepLearning, kendall2017WhatUncertaintiesWe}. 
The discussion about whether aleatoric uncertainty is \textit{irreducible} and epistemic \textit{reducible} resembles the dissent between frequentist and Bayesian views as well, since reducibility depends heavily on what is defined as knowledge.\phantomsection\label{irreducible}
At first, knowledge related in general to the ``behaviour of the system'' \citep{HORA1996217} and could include expert knowledge, additional data and variables, as well as knowledge about the correct model class and data generating process.
However, with the rise of deep neural networks and their flexibility, it is implicitly assumed that the hypothesis space (or model class) is large enough to capture the ground truth.
As a result, it is assumed that epistemic uncertainty relates only to finding the correct parameters and can be reduced with more training data \citep{kendall2017WhatUncertaintiesWe, hullermeier2021AleatoricEpistemicUncertainty}.

\citet{HORA1996217} argued already that a clear and unique separation between the two types of uncertainty is often not possible. A simple example might demonstrate this. 
\phantomsection\label{rolling.dice.example}Although rolling a fair dice is commonly used as an example of pure randomness and thus aleatoric uncertainty, it can also be seen as a purely physical process. Knowing the initial position, each rotation and movement of the dice, it is possible to predict exactly which number will be rolled (see Laplace's demon). Since extending this thought might lead to philosophical debates about determinism, it is important to set an appropriate framework. 

We, therefore, utilize basic classical probability theory to disentangle aleatoric and epistemic uncertainty. We pursue this in the framework of supervised machine learning and consider the setting where we have {\sl input variables} (features) denoted as $X \in {\cal X}$ and {\sl output variables} (responses) $Y \in {\cal Y}$. Adding to \citet{hullermeier2021AleatoricEpistemicUncertainty}, aleatoric uncertainty is then the uncertainty originating from a stochastic or non-deterministic relationship between the input variables $X$ taking value $x$ and the output $Y$, and can be described by a probability model. To be specific, we consider the conditional probability model
\begin{equation}
\label{eq:model0}
    F_{Y|X}(y|x) \coloneqq \mathbb{P}(Y \le y | x ) 
\end{equation}
where $F_{Y|X}$ denotes the conditional cumulative distribution function of $Y$ given input $x$. 
It is important to note that this probability-theoretic perspective is independent of the specific modelling approach or model selection process. 
Consequently, algorithms of any complexity—such as regression models, tree-based methods, or neural networks—can be employed to learn the prediction model or conditional relationship $Y|X$.
Following statistical notation, we rewrite (\ref{eq:model0}) as follows. First, for notational simplicity, we assume that the distribution function has a density (or probability function), which we denote as $f_{Y|X}$. If $F_{Y|X}(y|x)$ is differentiable for all $x \in {\cal X}$ and  $y \in {\cal Y}$, we have the well-known relation $f_{Y|X}(y|x) = (\partial F_{Y|X}(y|x)) / (\partial y) $. We now use notation $f_{Y|X}$ to write probability model (\ref{eq:model0}) as
\begin{equation}
\label{eq:aleatoric}
    Y|x \sim f_{Y|X}(\cdot|x).
\end{equation}
Unless stated otherwise, all our derivations also apply to categorical outcomes $Y$, in which case $f_{Y|X}$ is a probability mass function. For the sake of simplicity, we forego the use of formally precise notation from measure theory.

The probability model in equation (\ref{eq:model0}) and (\ref{eq:aleatoric}), respectively, serves as a definition for aleatoric uncertainty.
This equation states that this theoretical, \textit{true} relationship between input and output is probabilistic and thus described by a probability distribution. This distribution or probability model is typically unknown and we wish to learn it. For example, the underlying probability model may represent a physical process governed by the true physical laws of the universe or an economic system involving utility-maximizing agents. Crucially, (\ref{eq:model0}) and (\ref{eq:aleatoric}) are the complete, \textit{true} relationship between $Y$ and $X$, sometimes also called ``nature''. Thus, (\ref{eq:model0}) and (\ref{eq:aleatoric}) do not represent a specific machine learning model or model class such as a neural network (see~\ref{subsubsec:statistical.model.parametric.modell.class}). Also, (\ref{eq:model0}) and (\ref{eq:aleatoric}) are not tainted by the impurities of real data that have been collected (we will later discuss, e.g., measurement errors and missingness). Learning the truth, i.e., (\ref{eq:model0}) and (\ref{eq:aleatoric}), would be the \textit{non plus ultra} of knowledge gain.

%\cg{last sentence directly taken from Review2}
%\cg{add clarification that eq 1 is conditional density and then calculating the expectation would be e.g. prediction model}

Typically, supervised machine learning is interested in a prediction model for $Y$ given input value $x$. 
% Such a prediction model can be obtained by, e.g., minimizing the L2 loss. Denoting with $\hat{Y} = m(x)$ the prediction of $Y$ given $x$, then the best L2 prediction is obtained through minimizing the conditional mean squared error
% \begin{equation}
%     \operatornamewithlimits{argmin}_m E_{Y|X} \big( (Y - m(x))^2|x\big) 
% \end{equation}
% and it is easy to derive that the optimal prediction model results as
% \begin{equation}
%     m(x) = \mu(x) = E_{Y|X}\big(Y |x\big) 
% \end{equation}
% where $E_{Y|X}(\cdot | x)$ denotes the expectation, calculated  with respect to the conditional distribution $f_{Y|X}(\cdot|x)$.
As long as the conditional distribution (\ref{eq:aleatoric}) is not degenerated, we cannot predict $Y$ given $x$ without error. 
Non-degeneracy means here that the distribution is not concentrated to a single value $y_0$. Simply speaking and assuming that the variance exists, \textbf{aleatoric uncertainty} refers to a strictly positive variance ${Var}(Y|x)$. If ${\cal Y}$ is discrete-valued, this corresponds to a positive entropy of the discrete probability function (\ref{eq:aleatoric}). Therefore, all aleatoric uncertainty stems from the inherent stochasticity of the system. With this definition of aleatoric uncertainty, we can then also define \textbf{epistemic uncertainty} as all remaining uncertainty that arises in predictive modelling that is not aleatoric. 
Sometimes epistemic uncertainty is further divided into \textit{model uncertainty}, which relates to the correct specification of the structure of a model class, and \textit{parametric uncertainty} or \textit{estimation uncertainty}, which relates to the correct estimation of model parameters \citep{sullivan2015IntroductionUncertaintyQuantification}.
In the literature, the terms are not used consistently, as some authors use the term ``approximation uncertainty'' to describe estimation uncertainty \citep{hullermeier2021AleatoricEpistemicUncertainty}. However, since the ``approximation error'' in statistics and machine learning is commonly understood as the difference between the best model in the hypothesis space and the optimal model, it rather relates to model uncertainty \citep{mohri2018FoundationsMachineLearning}. To avoid further confusion, we stick to the term ``estimation uncertainty''.
We conclude with the following definition of aleatoric and epistemic uncertainty: Aleatoric uncertainty is defined as ${Var}(Y|x)$. All remaining uncertainty is defined as epistemic uncertainty.

With this definitional framework, we can look at the uncertainties in the dice example from earlier again. If no additional variables are gathered, it is not possible to further reduce the uncertainty concerning the fair dice roll and all uncertainty is of aleatoric nature. 
If however, all relevant physical quantities are measured (assuming they would be quantifiable), the process of rolling a dice becomes deterministic and no aleatoric uncertainty is present. Uncertainties can therefore only be distinguished on a model and data basis and not universally.
\phantomsection\label{uncertainties.only.based.on.a.model.not.universally} 
This context dependency of classifying uncertainties is an important takeaway and recurring theme throughout this paper. 

The success of describing the relationship between $Y$ and $X$ as stochastic as in \eqref{eq:aleatoric} does not come from showing that a deterministic view of the world might be incorrect. Rather, it stems from determinists and non-determinists agreeing that in the vast majority of practical applications, the data are such that $Y$ cannot be explained or predicted from $X$ without error and treating the relationship as stochastic is very useful: i.e., data analysis operates between the two extremes of the dice rolling example: no (useful) features or complete determinism. 
\subsection{Aleatoric and Epistemic Uncertainty in Classical Statistics}
\label{sec:statistics_and_uncertainty}

%\rtwo{The linear regression model is used to illustrate the key concepts. However, this does not help the reader to gain insights into more involved, yet simple models widely used, such as Gaussian process regression. Adding such a working example would be a step forward in partially justifying the title of the paper.}

\subsubsection{Statistical Model}\label{subsubsec:statistical.model.parametric.modell.class}
%\fk{the section above also started with reference to historic statistics, how is that different? anfänge der abschnitte, wie sind die einzuordnen, wo ist der rote faden, warum lese ich diesen teil, wieos ists relevant für gesamt story? -> todo abklären/angleichen.}
%\gk{Guter Punkt. Ich habe hier gekürzt und geschärft}

We further discuss the distinction between aleatoric and epistemic uncertainty 
using linear regression, which allows illustrating the complexity even in a simple setting.
Note that in general,  we do not know the aleatoric framework (\ref{eq:aleatoric}) but aim to estimate or learn a (prediction) model for output $Y$ given input $X$. In other words, for learning we replace (\ref{eq:aleatoric}) by
\begin{equation}
\label{eq:approx}
    Y |x \sim p_{Y|X}(\cdot|x)
\end{equation}
where $p_{Y|X}(\cdot)$ serves as an approximation for $f_{Y|X}(\cdot)$. Setting \eqref{eq:approx} can thereby be considered a statistical model, e.g., a regression model 
\citep{fahrmeir.et.al.2022.regression} or any kind of machine learning related model
such as a deep neural network \citep{goodfellow2016DeepLearning}. Note that in the terminology of machine learning the choice of model (\ref{eq:approx}) includes both, the choice of a loss function (which corresponds to choosing a specific probability model) as well as modelling how the parameters of the model depend on the input variables $X$. In the terminology of machine learning (\ref{eq:approx}) the latter is usually called hypothesis space 
\citep[p.~696]{russell2010ArtificialIntelligenceModern}, 
and one usually assumes some parametric model space  ${\cal P} = \{p_{Y|X}(y|x ; \theta): \theta \in \Theta\}$\label{eq:def.hypothesis.space.P}.  The parameters need to be set in some optimal way which we assume to be done based on training data ${\cal D} = \{ (x^{(i)}, y^{(i)}), i=1, \ldots, n\}$.
For now, we assume a simple $i.i.d.$ setting, i.e., we assume independence and that for given $x$ the response $y$ is generated from the true relationship (\ref{eq:aleatoric}). 
Based on the data ${\cal D}$  we obtain the trained model $\hat{p}_{Y|X}(\cdot|x)$, which in case of a parametric hypothesis space equals $\hat{p}_{Y|X}(\cdot|x) = p_{Y|X}(\cdot|x,\hat{\theta}) $ with $\hat{\theta}$ as fitted or learned parameter.

\begin{figure}[tb!]
    \centering
    \includegraphics[width= 0.9\linewidth]{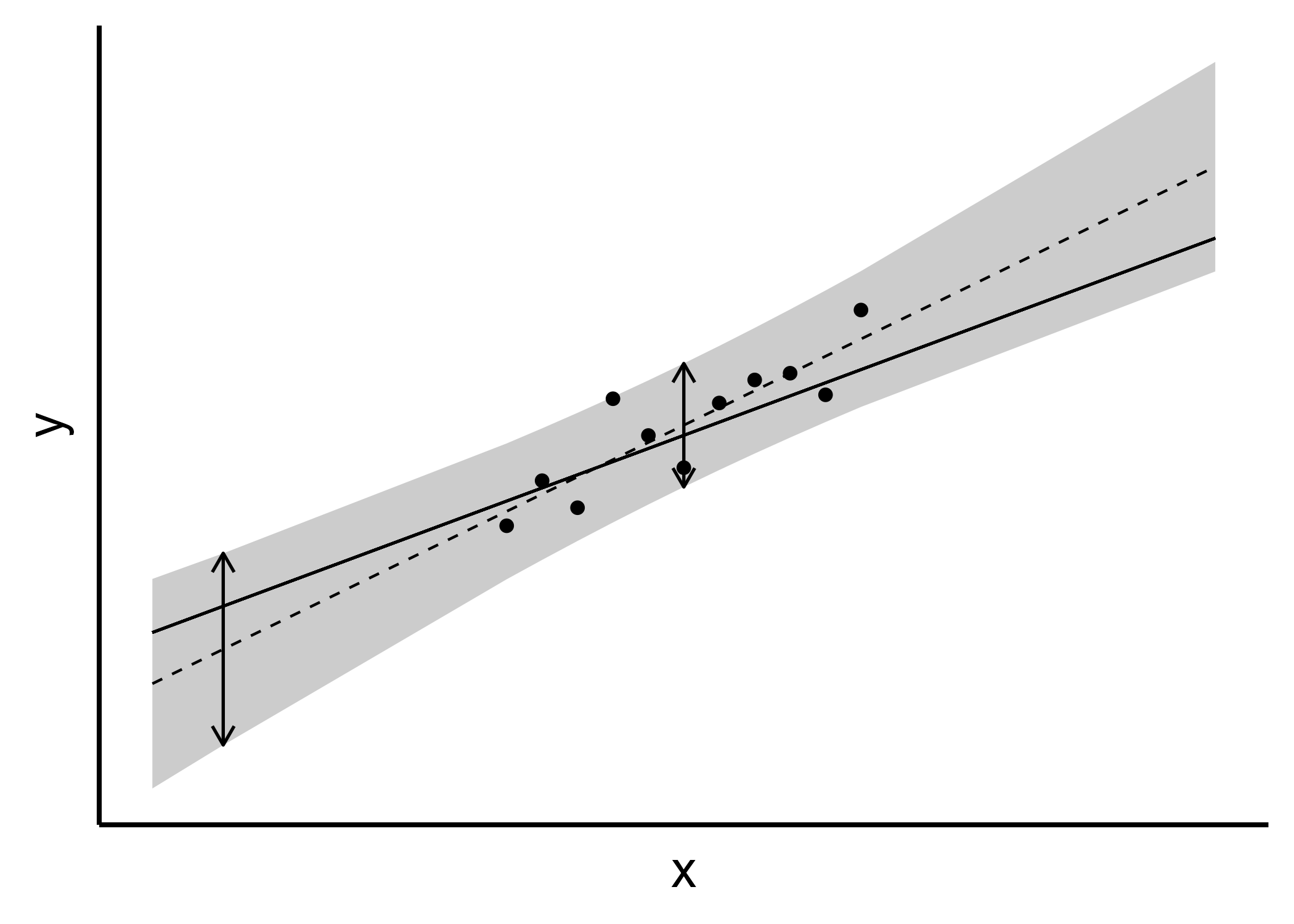}
    \caption{Linear model example. The solid line shows the true relationship between $x$ and $y$.  The estimated regression line is dashed. 90\% prediction interval in grey. Even though the true model class or hypothesis space is known, i.e., simple linear model (\textit{no model uncertainty}), it is not possible to predict $y$ precisely (\textit{aleatoric uncertainty}). Since the regression line is estimated with finite data, there is a discrepancy between the true parameters $\theta$ and the estimate $\hat{\theta}$ (\textit{estimation uncertainty}).}
    \label{fig:regression_uncertainty}
\end{figure}

\phantomsection
\label{regression.example.for.aleatoric.uncertainty}
A simple, though unrealistic setting is to assume $f_{Y|X}(y|x)$ in (\ref{eq:aleatoric}) is an element of ${\cal P}$, that is we know $f_{Y|X}(y|x) = p_{Y|X}(y|x, \theta_0)$ for some parameter  $\theta_0$ which is then usually called the \textit{true} parameter. This means the model is well-specified (and not misspecified) and no model uncertainty is present. 
We consider this setup in a simple linear model where $X$ and $Y$ are related through a linear relationship with residual noise, i.e., $Y=\beta_0 + x \beta_{1} + \epsilon$ with $\epsilon \sim N(0, \sigma^2)$. In other words, we assume 
\begin{align}
    \label{eq:linmod}
    Y|x \sim N(\beta_0 + x \beta_x , \sigma^2)
\end{align}
The general aim is to estimate parameters $\theta = (\beta_0, \beta_1, \sigma^2)$ based on some data $(x^{(1)}, y^{(1)}), \ldots, (x^{(n)}, y^{(n)})$. As model \eqref{eq:linmod} is learned from a finite amount of data, it is in general not possible to recover the true parameters $\theta$.
Letting $\hat{\theta}$ denote the parameter estimate based on the data, we can quantify the discrepancy between $\hat{\theta}$ and $\theta$, which in statistics is also commonly called \textit{estimation uncertainty}, expressed in \textit{estimation variance}, which reduces with increasing the amount of data. 
Still, \textit{aleatoric uncertainty} remains due to the variance of the residuals $\epsilon$, which becomes apparent when calculating prediction intervals. 
The prediction interval for a future observation $y_0$ at location $x_0$ with level $1 - \alpha$, i.e., a measure of the total uncertainty, is given by
\begin{align}
   {x}_0^{\prime} \hat{{\beta}} \pm t_{(n-p), (1-\alpha / 2)} \hat{\sigma}\left(1+{x}_0^{\prime}\left({X}^{\prime} {X}\right)^{-1} {x}_0\right)^{1 / 2}
   \label{eq:prediction.interval.linear.regression}
\end{align}
\citep{fahrmeir.et.al.2022.regression}, consisting of both aleatoric and estimation uncertainty. We visualize this in Figure~\ref{fig:regression_uncertainty}.
It is important to note that even in this very simple model one cannot additively decompose the total uncertainty into aleatoric and estimation uncertainty. This means, while conceptually uncertainty consists of aleatoric and epistemic uncertainty, this is not mirrored mathematically in the form of added noise. 

Furthermore, in many applications, we will have a misspecified model, i.e., $f_{Y|X} \not\in {\cal P}$. 
Besides the already mentioned uncertainties, \textit{model uncertainty} or \textit{model bias} exists in this case as well. This is caused by the discrepancy between $f_{Y|X}$ and $p_{Y|X}$, i.e., since in general \eqref{eq:approx} is only approximating \eqref{eq:aleatoric}.
In other words, the complexity increases readily, if the model class is only an approximation of the true underlying data generating process \eqref{eq:aleatoric}, which is in real-world data more or less always the case. 

%\cg{Add example with gaussian process? is it possible to distinguish AU, EU?}

We can also take Gaussian processes as an example to disentangle aleatoric and epistemic uncertainty. We assume in this case that $Y|x$ is normally distributed with 
\begin{align}
    \label{eq:gp1}
    Y|x \sim N( \mu(x) , \sigma^2(x))
\end{align}
where $\mu(x)$ is the conditional expectation of $Y$ given $x$ and $\sigma^2(x)$ the conditional variance. Assuming for simplicity that $\sigma^2(x) $ is constant and does not depend on $x$, which is called variance homogeneity in statistical jargon, we can state that aleatoric uncertainty corresponds to positive variance $\sigma^2$. Given that $\mu(x)$ is unknown, one can assume that $\mu(x)$ follows a Gaussian process with mean value $0$ and covariance structure 
$Cov(\mu(x), \mu(x'))$ which is also called kernel and typically modelled as some decreasing function in the distance between $x$ and $x'$, see e.g., \citet{seeger2004gaussian} or \citet{rasmussen2006GaussianProcessesMachine}. The uncertainty about the true $\mu(x)$ and $\sigma^2$ is part of the epistemic uncertainty. Again, a decomposition of aleatoric and epistemic uncertainty is only possible conceptually.

Recently, there has been increasing interest in Gaussian process methods in machine learning \citep{rasmussen2006GaussianProcessesMachine, liu2020WhenGaussianProcess, wang2022GaussianProcessRegression}. In spatial analyses such as in geostatistics, Gaussian processes are known as Kriging \citep{cressie1990OriginsKriging}.

\subsubsection{Aleatoric and Epistemic Uncertainty in the Bias-Variance Decomposition}
The parallels in the linear model of residual variance to aleatoric uncertainty on the one hand, and model bias and estimation variance to epistemic uncertainty on the other hand, can also be mirrored in the classical bias-variance decomposition.
The bias-variance decomposition shows that the expected prediction error can be decomposed into the \textit{model's bias}, \textit{variance}, and an \textit{irreducible error} (\citeauthor{geman1992neural}, \citeyear{geman1992neural}; \citeauthor{hastie2009ElementsStatisticalLearning}, \citeyear{hastie2009ElementsStatisticalLearning}, p.~223; \citeauthor{rossetFixedXRandomXRegression2020}, \citeyear{rossetFixedXRandomXRegression2020}).
Let $ \hat{y} = \hat{\mu}(x) = \int y \: \hat{p}_{Y|X}(y|x) dy$ be the mean prediction, given $x$, based on the fitted model. Then the well-known bias-variance decomposition of the mean squared prediction error results:
 \begin{align}
 \label{eq:decomposition.of.prediction.error}
     E_{Y|X}&\{ E_{\cal D} \{ ( Y - \hat{y})^2\} |x\}  =  \\
     &E_{Y|X}\{ E_{\cal D}  \{ ( Y - E_{Y|X}(Y|x) + E_{Y|X}(Y|x) - \hat{y})^2 \} |x  \} \nonumber  \\ %\displaybreak %if needed for alignment on page
      =& {Var}_{Y|X}(Y|x)  + \nonumber \\
      & E_{\cal D}\{  ( \hat{y}  - E_{\cal D}(\hat{y}|x) +  E_{\cal D}(\hat{y}|x)  - E_{Y|X}(Y|x) )^2 \} \nonumber \\
      = &  {Var}_{Y|X}(Y|x) + {Var}_{\cal D}(\hat{y}|x) + {bias}( \hat{y}|x )^2 \nonumber
 \end{align}
where $  bias( \hat{y} |x) = E_{\cal D}(\hat{y}|x) - E_{Y|X}(Y|x)$ is the bias in the conditional mean prediction. With $E_{\cal D}()$ we thereby refer to the expectation over the data ${\cal D}$.
Here the direct link between aleatoric and epistemic uncertainty and the classical bias-variance trade-off becomes clear. ${Var}_{Y|X}(Y|x)$ corresponds to aleatoric uncertainty and cannot be reduced and the model $\hat{p}_{Y|X}$ does not figure into this term at all. The variance of the estimator $\hat{y}$ corresponds to estimation uncertainty, while the bias of the estimator relates to model uncertainty.
Both ${Var}_{\cal D}(\hat{y}|x) \text{ and } {bias}( \hat{y} |x)^2$ depend on the model's complexity, as a more flexible model typically induces a lower bias but higher variance. 
Therefore, an optimal model complexity exists that balances bias and variance and thus leads to the lowest possible generalization error.

According to one of the most-cited textbooks on statistical learning, ``it is generally not possible to explicitly compute the test MSE, bias, or variance for a statistical learning method'' from real-world datasets, since the expectation $E_{Y|X}(Y|x)$ of the true model in \eqref{eq:decomposition.of.prediction.error} is not known \cite[p. 36]{james2023IntroductionStatisticalLearning}. Yet, in more thorough discussions the same author and others \citep{jamesVarianceBiasGeneral2003, sallesBiasvarianceAnalysisStateoftheart2021} try to compute exactly this, except that their analysis is based on 0-1 loss and for categorical outcomes, by approximating the aleatoric uncertainty in a first step. The aforementioned literature shows, again, the challenges involved if one tries to decompose total uncertainty into aleatoric and epistemic uncertainty on an empirical basis.

The bias-variance decomposition is well known for models where the number of parameters is less than $n$, which in particular in deep neural networks does not usually hold \citep[see][]{belkin2019ReconcilingModernMachinelearning, clarte2022StudyUncertaintyQuantification}.
Recent literature focuses on model performance if in contrast $p > n$,
see for instance \citet{efron2020prediction},
\citet{fan2021selective}, \citet{bartlett2021deep} or \citet{hastie2022surprises}, to name but a few.
In fact, the success of deep learning is based on heavily overparameterized models which utilize the fact that a second bias-variance trade-off optimum, can occur in constellations with more parameters than observations. This phenomenon is also known as \textit{double descent} \citep{belkin2019ReconcilingModernMachinelearning, neal2018ModernTakeBiasVariance, neyshabur2018UnderstandingRoleOverParametrization, yang2020RethinkingBiasvarianceTradeoff, zhang2021UnderstandingDeepLearning}. 
Double descent is neither fully understood yet, nor is its connection to uncertainty fully explored.
Thus, we do not want to get deeper into the double descent discussion in this paper, as our focus lies on the source of uncertainty and not on improving prediction error. We do consider the topic very relevant though and additionally refer to \citet{nakkiran2021DeepDoubleDescent}, \citet{clarte2022StudyUncertaintyQuantification} or \citet{mei2022generalization}.

%Instead, a phenomenon is observed, where an even higher model complexity, exceeding the interpolation threshold, i.e., $n \ll p$ leads to a low generalization error, again. This second descent in the error curve gave the phenomenon its name ``double descent'' \citep{nakkiran2021DeepDoubleDescent}.

%This is focused on in the next section. 
%\input{3-Sources/overparam-short}

\subsection{Kullback-Leibler Divergence and Misspecified Models}
\label{sec:overparameterized}
For dealing with model uncertainty, we want to look more closely on the role of parameter estimation in case of misspecified models. To start with, let again ${\cal P} = \{ p_{Y|X}(y|x; \theta): \theta \in \Theta\}$ be the hypothesis space, that is the set of possible prediction models. We assume $\Theta \subset \mathbb{R}^p$, i.e.,\ we consider a $p$-dimensional parameter space where for the moment we assume the classical statistics framework $p<n$. We make use of $p(y|x;\theta) \in {\cal P}$ as an approximate model for $f(y|x)$ and measure the ``distance'' between $f(\cdot)$ and $p(\cdot)$ through the Kullback-Leibler divergence. 
The optimal parameter $\theta_0$ can then be defined by minimizing the (expected) Kullback-Leibler loss, that is
\begin{align}
\label{eq:theta0}
    \theta_0 &= \underset{\theta \in \Theta}{\mbox{arg min}} \; E_X\left\{KL(f(\cdot|X), p(\cdot|X;\theta))\right\} \\ 
    \nonumber
    &= \underset{\theta \in \Theta}{\mbox{arg min}}\int_{x \in {\cal X}} \int_{y \in {\cal Y}}  \log \left( \frac{f(y|x)}{p(y|x;\theta)} \right) f(y|x) f(x) dy \,  dx \\ 
    &= \underset{\theta \in \Theta}{\mbox{arg max}} \int_{x \in {\cal X}} \int_{y \in {\cal Y}}  \log p(y|x;\theta) f(y|x) f(x)  dy \, dx \nonumber
\end{align}
where we assume for simplicity of notation that both, $y$ and $x$ are continuous. For our approximate model $p(y|x,\theta)$ we assume in the following Fisher regularity (see \citealp{schervish2012theory} or \citealp{kauermann2021statistical}). Particularly, this means that differentiation with respect to the parameter and integration with respect to $y$ (and $x$) are exchangeable, which in particular requires that the support of $Y$ does not depend on the parameter and that the parameter space $\Theta$ is open. 
Then, differentiation defines $\theta_0$ implicitly through 
\begin{align}
\label{eq:theta0-1}
    0 = \int_{x \in {\cal X}} \int_{y \in {\cal Y}} \frac{\partial \log p(y|x;\theta_0)}{\partial\theta} f(y|x) f(x) dy \, dx.
\end{align}
%[check whether we can pull $\partial/\partial \theta$ out]
Taking the
%and it can be shown that the maximum likelihood estimate $\hat{\theta}$ converges to $\theta_0$, where $\hat{\theta}$ maximizes the
log-likelihood function
\begin{align}
    \label{eq:loglik}
    l(\theta)= \sum_{i=1}^n \log p(y^{(i)}|x^{(i)};\theta)
\end{align}
we can argue that based on the law of large numbers $l(\theta) / n $ is a consistent estimate of the integral in (\ref{eq:theta0}). 
We obtain the maximum likelihood estimate $\hat{\theta}$ implicitly through 
\begin{align}
    \label{eq:loglik2}
    0 = \frac{1}{n}\sum_{i=1}^n \frac{\partial \log p(y^{(i)}|x^{(i)};\hat{\theta})}{\partial \theta}
\end{align}
and (\ref{eq:loglik2}) can be seen as the empirical version to (\ref{eq:theta0-1}).
If the estimate derived from (\ref{eq:loglik2}) exists, that is if the likelihood function (\ref{eq:loglik}) is convex (and the maximum is not at the boundary of $\Theta$), it can be shown that the maximum likelihood estimate $\hat{\theta}$ converges to $\theta_0$. A sketch of the proof is provided in \autoref{appendix:convergence_MLE}, see also \citet[][p. 234]{kauermann2021statistical}. Note that this rationale does not depend on knowing the true model $f(\cdot)$, or $f(\cdot)$ being element of ${\cal P}$. We can therefore conclude that using the likelihood approach is advisable even in the case of misspecified models.

The Kullback-Leibler divergence can be estimated, at least up to an additive constant. In fact, the Akaike Information Criterion (AIC) \citep{akaike1998information}
\begin{align*}
   \mbox{AIC}(p(\cdot|x;\hat{\theta})) = 
   - 2 \sum_{i=1}^n \log p(y^{(i)}|x^{(i)}; \hat{\theta}) + 2 p
\end{align*}
can serve as an estimate of two times the (expected) Kullback-Leibler divergence shifted by the inestimable entropy 
\begin{align*}
    \int_{x \in {\cal X}} \int_{y \in {\cal Y}}   \log(f(y|x)) f(y|x) f(x) dy \, dx.
\end{align*}
The term $+2p$ serves to avoid model over-specification bias, although it is only an approximation based on knowing the true model.
See for instance \citet[p.~110]{wood2017GeneralizedAdditiveModels}, \citet[p.~150]{davison2003StatisticalModels}, or \citet[p.~236]{kauermann2021statistical}. 

Thus, the AIC enables choosing the optimal model, approximately even for misspecified models, i.e., the estimation uncertainty is minimized, even when high model uncertainty is present.

The above statements on statistical reasoning rely on  $p < n$ with $n$ as the sample size. 
Similar to the aforementioned double descent phenomenon, the relationship between model and estimation uncertainty in the overparameterized regime, i.e., when $p > n$ requires further research. 
\section{The Role of Data - Model Uncertainty Revisited}
\label{sec:role.of.data}

\subsection{General Comments}

Our focus in this section is the model space or hypothesis space ${\cal P}$, from which $\hat{p}_{Y|X}(\cdot|x)$ is trained. The choice of an appropriate model space ${\cal P}$ is important: only if the true probability model $f_{Y|X}(\cdot|x)$ given in (\ref{eq:aleatoric}) is an element of ${\cal P}$ can we hope to approximate the true model during training. Unfortunately, the true model is rarely known. As mentioned above, model uncertainty refers to the unknown discrepancies between the true model and the optimal model in ${\cal P}$, discrepancies that should be reduced as much as possible. The high capacity of machine learning models to learn very flexible functions are one way of doing this. Often, aleatoric uncertainty is neglected or simplified.
Besides, data for training models are often deficient, leading analysts to end up with suboptimal models, often even without considering potentially superior alternatives. We, therefore, want to illuminate the role of data.
In this section, we provide examples of misspecified models in machine learning, in which the (arguably) true model is not part of the model space ${\cal P}$. The respective misspecifications are caused by shortcomings of the training data, particularly the absence of some variables. While this situation cannot always be avoided, our discussion serves as a reminder of the consequences: predictions can be biased and variances may increase or even be underestimated, contingent on the choice of ${\cal P}$. This is evidence that a perfect separation between model uncertainty and aleatoric uncertainty is not possible.
\subsection{Uncertainty Due to Unobserved Variables}
\label{subsec:biases.due.to.unobserved.variables}
\subsubsection{General Framework}   \label{subsubsec:biases.due.to.unobserved.variables.general.framework}

We consider three groups of quantities: $X$, $Y$, and $Z$. As above $X \in {\cal X}$  are {\sl input variables} and $Y \in {\cal Y}$ are  {\sl output variables}. We additionally introduce unrecorded or {\sl unobserved variables} $Z \in{\cal Z}$. These variables may relate to $X$, to $Y$, or to both. The unobserved variables $Z$ can also relate to how the data are collected. Moreover,
$Z$ can contain parts that were not measured in the given data but are quantifiable in principle, as well as parts that are always unobservable or are infeasible to observe with acceptable efforts. 
The key point is that there may be unmeasured context information $Z$ that influences the value of $Y$ or $X$ or both. We will motivate the role of $Z$ with a number of examples throughout the rest of the section. 

We assume that the triple $(X,Y,Z)$ is drawn from some superpopulation according to 
\begin{align}
%\label{eq:standarddgp}
\label{eq:true_xyz}
      & (X,Y,Z)  \sim  f_{X,Y,Z}(x,y,z)  \\
      &= f_{Y| X,Z}(y|x,z) \: f_{X|Z}(x|z) \: f_Z(z)     \nonumber 
\end{align}
where the factorization follows from the chain rule (or general product rule) of probability theory and thus always holds. However, the separate components in the joint distribution $f_{X,Y,Z}$  are typically not known in real-world applications. 

Available training data do not include the unobserved variables $Z$. We only have access to the observed variables $X$ and $Y$ in the database. By applying the law of total probability to \eqref{eq:true_xyz}, we obtain the joint distribution of $(Y, X)$ as
\begin{align}
    \label{eq:datagen} 
    &(Y, X) \sim f_{Y, X}(y, x)  \\
    &= \int_{z \in {\cal Z}}  f_{Y|X,Z} (y|x,z) f_{X|Z} (x|z) f_{Z}(z) dz. \nonumber
\end{align}
We may also generally condition equation \eqref{eq:datagen} on $x$ and obtain 
\begin{align}
\label{eq:datpred}
  &(Y |x) \sim f_{Y|X}(y|x)    \\
    &= \int_{z \in {\cal Z}} f_{Y|X,Z}(y|x,z) \: f_{Z|X}(z|x) dz. \nonumber
\end{align}
The above integrals assume a suitable measure for integration, i.e., if $Z$ is discrete-valued, the integral refers to summation, otherwise appropriate integration is assumed.

In supervised machine learning, we wish to learn the conditional distribution $f_{Y|X}(\cdot | x)$.\phantomsection\label{true.relationship.as.reference.point} We may consider this relationship  $f_{Y|X}(\cdot | x)$ as the best any model can achieve. \phantomsection\label{best.a.model.can.do}
It is thus the natural reference point for our subsequent analyses in which data deficiencies are the cause of why this optimum may not be reached and consequently any data-driven model can be biased and/or carries additional uncertainty.

Often, but not always, we are interested in a prediction of $Y$ given $x$. In this case the conditional mean value
\begin{align}
\label{eq:condmean}
    \mu(x) \coloneqq E_{Y|X}(Y|x) = \int_{y \in {\cal Y}} y f_{Y|X}(y|x) dy
\end{align}
serves as best prediction, based on minimizing the squared prediction error.
\subsubsection{Omitted Variables}   
\label{subsec:omitted.variables}
\label{subsubsec:omitted.variables}

Even though omitted variables or missing features did not get much attention in machine learning so far, they are a relevant source of a model's uncertainty (see \citealp{chernozhukov2021omitted}). 
Not only in structured, tabular data, but also in unstructured data, problems can arise if important features are ignored or not collected. 
Examples include missing colour information in black and white images or missing wavelengths or sensors such as radar or lidar which could be relevant in autonomous driving or remote sensing \citep{li2020LidarAutonomousDriving, zhu2017DeepLearningRemote}.
In the example of rolling a dice (see~\ref{rolling.dice.example}), any uncertainty was due to our ignorance of physical processes. Similarly, the data recorded by mobile health devices depend on the physical state of a person, which is not logged. 
Formalized in this framework, omitted variables are features $Z$ that affect outcome $Y$ and possibly $X$, but that are not included in $X$ as they have not been observed.

The comparison is between $Y|x,z$, the true relationship which we will also call the ``full model'', and $Y|x$, the ``omitted variable model''. We assume that we want to predict the conditional mean value of $Y$. From \eqref{eq:datpred} it follows that
\begin{align}
    \label{eq:omitted.expected.value}
    E_{Y|X}(Y|x) & = \int_{y \in {\cal Y}} \int_{z \in {\cal Z}}  y \: f_{Y|X,Z}(y|x,z)  f_{Z|X}(z|x) \: dz \: dy \\
    & = \int_{z \in {\cal Z}} E_{Y|X,Z}(Y | x, z) f_{Z|X}(z|x) \: dz, \nonumber
\end{align}
implying that predictions $E_{Y|X,Z}(Y | x, z)$ based on $x$ and $z$ would typically be more precise than their (weighted) average $E_{Y|X}(Y|x)$. For given $x$ and $z$, the difference $E_{Y|X,Z}(Y|x,z) - E_{Y|X}(Y|x) \eqqcolon bias(x,z)$ is the bias in the conditional prediction of $Y$ due to the omission of $Z$. In order to assess (aleatoric) uncertainty, we calculate the conditional variance in the omitted variable model using the law of total variance, conditional on $X=x$,  
\begin{align}
\label{eq:varincrease}
    Var&_{Y|X}(Y|x)  =   
  \int_{z \in {\cal Z}}  Var_{Y|X,Z}(Y| x, z)   f_{Z|X}(z|x) \: dz \\
  & +  \int_{z \in {\cal Z}}
  \left(E_{Y|X,Z}(Y|x,z) - E_{Y|X}(Y|x)\right)^2\notag \\
  &\cdot f_{Z|X}(z|x) dz        \notag \\
   =& E_{Z|X}\left( Var_{Y|X,Z}(Y| x, Z)\right) +  E_{Z|X}\left( bias(x,Z)^2 \right). \nonumber
\end{align}
We start by noting that the more predictive the omitted variable $Z$, the larger the expected squared bias can be. From the second term on the right-hand side of \eqref{eq:varincrease} we find that, \textit{ceteris paribus},
the larger the expected squared bias, the larger the conditional variance of $Y$ is in the omitted variable model relative to that in the full model. This reflects common notions of aleatoric uncertainty and omitted variables: if one had additional features $Z$, one could predict $Y$ with less uncertainty. 

However, this average or marginal view hides important aspects when $ Var_{Y|X,Z}(Y| x, z)$ or $ bias(x, z)$ are not constant for all $z$. Therefore, to contrast the uncertainty between the full and the omitted variable model, we compare the conditional variances $Var_{Y|X,Z}(Y|x,z)$ and $Var_{Y|X}(Y|x)$. 
As we show in Appendix~\ref{sec:app.omitted variables}, %{subsubsec:app.omitted variables.math.compare.variances}, 
there always exist $z \in {\cal Z}$ for which 
%\[
$Var_{Y|X,Z}(Y|x,z)  \le  Var_{Y|X}(Y|x)$ 
%\]
and, unless $Z$ affects neither variance nor expected value, we even have $Var_{Y|X,Z}(Y|x,z) < Var_{Y|X}(Y|x)$. This is still in line with the aforementioned common notions. 
However, what may be less obvious is that there can also exist $z$ for which $Var_{Y|X,Z}(Y|x,z) > Var_{Y|X}(Y|x)$: that is,
the measure of aleatoric uncertainty employed in the omitted variable model can also be underestimated for some data points. Therefore, even when there is no bias, prediction intervals can suffer at the same time from both, over-coverage and under-coverage, depending on $z$. We also note that the quantification of epistemic uncertainty is affected when it is defined as the difference between total and aleatoric uncertainty (and total uncertainty is measured correctly). 

Consider a scenario in which $Z$ is not predictive of $Y$ in any way. Then, in \eqref{eq:varincrease} the expected squared bias is zero but $Var_{Y|X,Z}(Y|x,z)$ can still depend on $z$, with the consequences discussed in the previous paragraph. Even if such a feature $Z$ was available in the data, it would not get selected into the prediction model during training using typical loss functions. %is purely the prediction error. 
It is not common in the typical machine learning workflow to consider a non-predictive $Z$, even though it may be relevant for uncertainty quantification.
%In traditional statistical modelling, a  dedicated statistician or a domain expert with knowledge  about 
The role of $Z$ might however be checked for heteroscedasticity. 
One possible cause is that differently reliable annotators might have labelled the data. This is an example of connections between seemingly unrelated sources of uncertainty: here, omitted variables (e.g., the identity $Z$ of the annotators) and errors in data (see~\ref{subsubsec:errors.in.x} and~\ref{subsubsec:errors.in.y}). 

Finally, the comparison of ${Var_{Y|X,Z}(Y|x,z)}$ and \newline  $Var_{Y|X}(Y|x)$ evidences what we mentioned in the dice rolling example (see \ref{rolling.dice.example}): what one considers to be aleatoric uncertainty depends on the specific context -- here, which features are considered, i.e.,  $X$, and which are not, i.e., $Z$. In particular, from the viewpoint of the full model replacing the aleatoric uncertainty $Var_{Y|X,Z}(Y|x,z)$ with its expected value in \eqref{eq:varincrease} should still only describe aleatoric uncertainty. However, the expected squared bias would be regarded as epistemic, not aleatoric. Therefore, a clear decomposition of uncertainty is again questionable.

\phantomsection\label{omitted.variables.binary.Y.var.and.bias.are.linked} 
In general, either of the two terms on the right-hand side of \eqref{eq:varincrease} independently of the other can affect the conditional variance of the omitted variable model. 
However, for binary $Y$ there is only one ``parameter'', the conditional probability given $x$ and $z$, determining both, $Var_{Y|X,Z}(Y| x, z)$ and $E_{Y|X,Z}(Y|x,z)$. 
Hence, the two terms in \eqref{eq:varincrease} are connected: generally, the omitted variable model suffers either from both, neglected variance heterogeneity and bias for some $z$, or from neither \citep[see Supplementary Material][]{supp}.
This, in turn, casts more doubt on the clear distinction between and separability of aleatoric uncertainty and epistemic uncertainty.

The focus of this paper is on prediction and uncertainty and not on (the bias of parametric estimates of) marginal effects. Hence, the aforementioned existence of bias in the conditional predictions suffices for us. As this bias can occur even when $X$ and $Z$ are independent, we conclude this section by highlighting that omitted variables can be an important, non-ignorable source of uncertainty.
In statistics and related disciplines such as econometrics, there is a much larger body of literature about omitted variable bias and Simpson's paradox (e.g., \citealt{wagner1982simpson}) than in machine learning (but see, e.g., \citealt{sharma2022detecting}). This statistical literature typically focuses on specific, traditional model classes such as linear regression. Most of its results can also be shown for more general relationships $f_{Y|X, Z}$ with our framework: e.g., that the trained model attributes the marginal effects of the missing variable $Z$ to the included $X$ to the extent the two depend on each other \citep[see Supplementary Material][]{supp}.
\subsubsection{Errors in X}%: Measurement Error
\label{subsubsec:errors.in.x}

Imperfect measurement instruments, usage of proxy variables, and subjectivity in labelling decisions are among the sources of errors in data. Let $Z$ be the error-free version of features $X$. 
Such errors in X have already been identified as causes for increased uncertainty in neural network outputs in \citet{wright1999BayesianApproachNeuralnetwork}, where the noiseless input is represented as a latent variable, which is directly in line with our suggestion to use $Z$.
For Gaussian processes, modelling the input noise is straightforward and more realistic uncertainty quantification is possible if desired \citep{girard2002GaussianProcessPriors, mchutchon2011GaussianProcessTraining, 
rasmussen2006GaussianProcessesMachine}.
The main intuition about the consequences of such errors is that the effects of features with measurement error are partly attributed to features with no or with lower measurement error. 
Consider the following example in which $Z$ is a scenery of which an image $X$ is taken. While $Z$ is clear and without error, an image $X$ is taken with resolution error. We may also think of $Z$ as a high-resolution image of which a noisy version $X$ is taken. 
A further example is text data $Z$ that are quantified through word embeddings leading to $X$.
%Another example results by considering $Z$ as text data which are quantified through word-embedding leading to $X$.

\begin{figure}[tb]
    \centering
    \includegraphics[width= 0.8\linewidth]{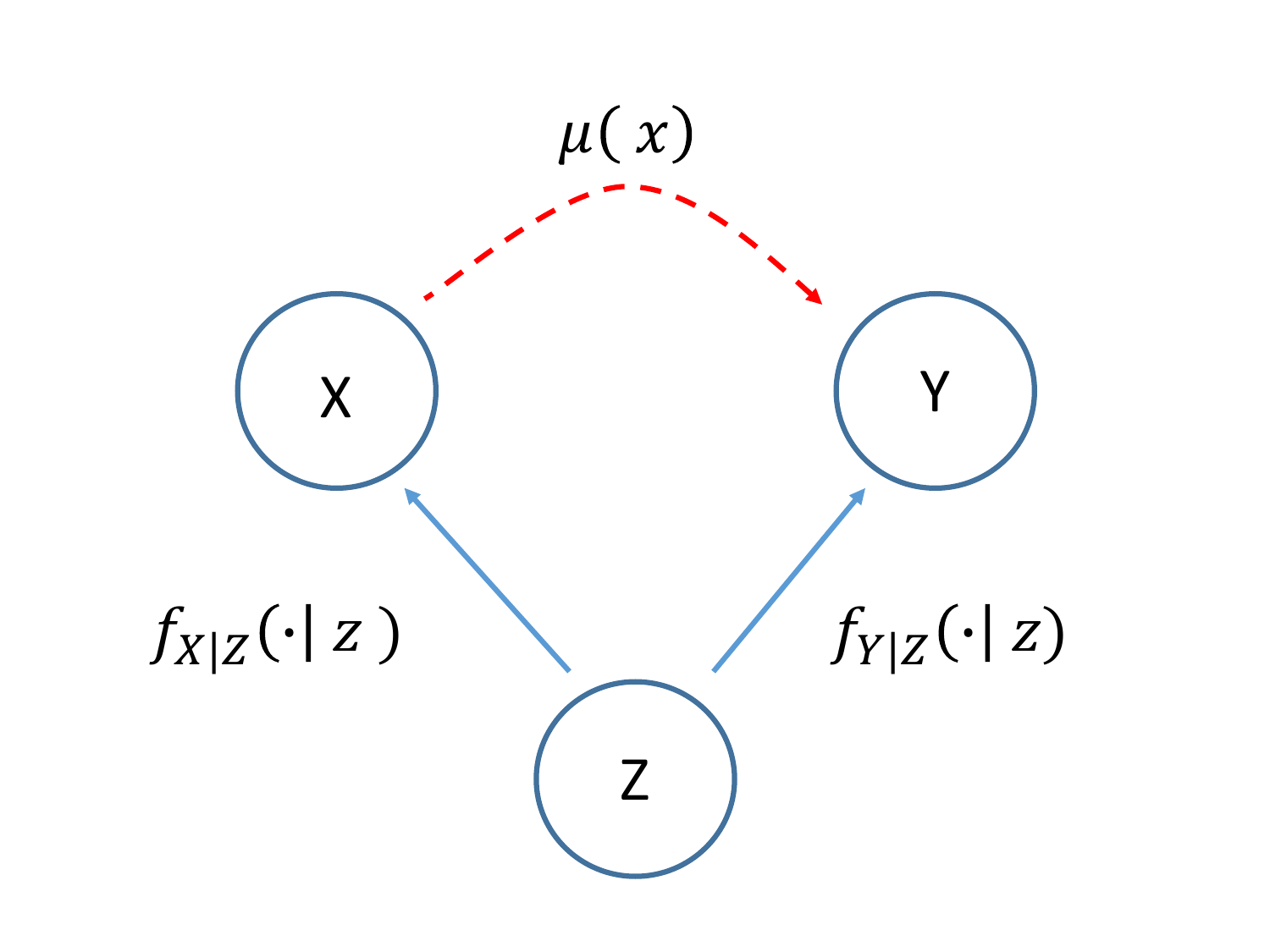}
    \caption{Errors in $X$ setting} 
    \label{fig:errorx}
\end{figure}

We consider the sources of uncertainty which come into play when $X$ is used instead of $Z$ to predict $Y$, as outlined in \autoref{fig:errorx}.
The exact effect on uncertainty depends on the relationships among the features $Z$ and of the features to the outcome $Y$, as well as on relationships concerning the errors (the errors among themselves, to the features, and to the outcome). This mirrors the previous subsection where the effects of omitted variables were partly attributed to the other features. In fact, the omitted variables situation is a limiting case: as the amount of measurement error in a mismeasured feature grows, the signal-to-noise ratio of this feature to the outcome decreases, and the effect of the feature is increasingly picked up by other features to the extent that they are related. In the limit, i.e., as the signal-to-noise ratio approaches zero, the affected feature may as well be omitted or is automatically omitted during model training.

The traditional errors-in-variables literature has focused on regression models (e.g, \citealp{carroll.et.al.2006.measurement.error}) and shows how parameter estimates are biased if covariates $X$ are measured with error. Several bias correction methods have been proposed, including simulation-based approaches \citep{lederer2006shortor} and methods using calibration data or repeated measurements. 
In our general framework \eqref{eq:true_xyz} - \eqref{eq:datpred}, 
the error mechanism is captured by $f_{X|Z}(\cdot|z)$ which links the observed, error-prone features $X$ to the unobserved, error-free values $Z$.  Following the example with high-resolution image $Z$ and its low-resolution version $X$, it is plausible to assume that $Y$ is independent of $X$ given $Z$, as sketched in Figure~\ref{fig:errorx} and discussed in Appendix~\ref{sec:app.errors.in.x.discussion.independence}. We notate this as
\begin{align}
\label{eq:condind}
    Y \perp\!\!\!\perp X | Z.
\end{align}
In this case, we may use 
\begin{align}
    \label{eq:errormodelvar}
Var_{Y|Z}(Y | z) = \int_{y \in {\cal Y}} (y - E(Y|z))^2 f_{Y|Z}(y|z) dy    
\end{align}
as a measure for the  ``error-free'' aleatoric uncertainty given $z$. Note that $Z$ remains unobservable and instead we observe the error-prone version $X$. 
This leads to the ``error-prone'' aleatoric uncertainty
\begin{align}
\label{eq:errormodel2}
    Var_{Y|X}&(Y|x) \\
 =& E_{Z|X}\{ Var_{Y|X,Z}(Y|x,Z)\}  \nonumber \\
    & + Var_{Z|X}\{E_{Y|X,Z}(Y|x,Z)\} 
    \nonumber \\
    =& E_{Z|X}\{Var_{Y|Z}(Y|Z)|x\} \nonumber \\
    &+ Var_{Z|X}\{E_{Y|Z}(Y|Z)|x\}, \nonumber 
\end{align}
which results by the law of total variation after conditioning on $X=x$ and utilizing (\ref{eq:condind}).
The first term is the average of component (\ref{eq:errormodelvar}) after observing $x$. If this depends only weakly on $x$, the first component in (\ref{eq:errormodel2}) is approximately equal to (\ref{eq:errormodelvar}). 
The second term in (\ref{eq:errormodel2}) leads to an increase in aleatoric uncertainty. 

Traditional statistical analysis tries to recover 
$f_{Y|Z}$ despite only observing $(Y,X)$. In predictive modelling, this is mainly of interest when $X$ is available in the training data but $Z$ in deployment. However, then the aleatoric and epistemic uncertainties of the correction model need to be considered, too. Interestingly enough, the bias-variance trade-off for how much correction is optimal for prediction has hardly been explored 
\citep[Ch.~3.5]{carroll.et.al.2006.measurement.error}.

\subsubsection{Errors in Y}     
\label{subsec:errors.in.y}\label{subsubsec:errors.in.y}
\begin{figure}[tb]
    \centering
    \includegraphics[width= 0.8\linewidth]{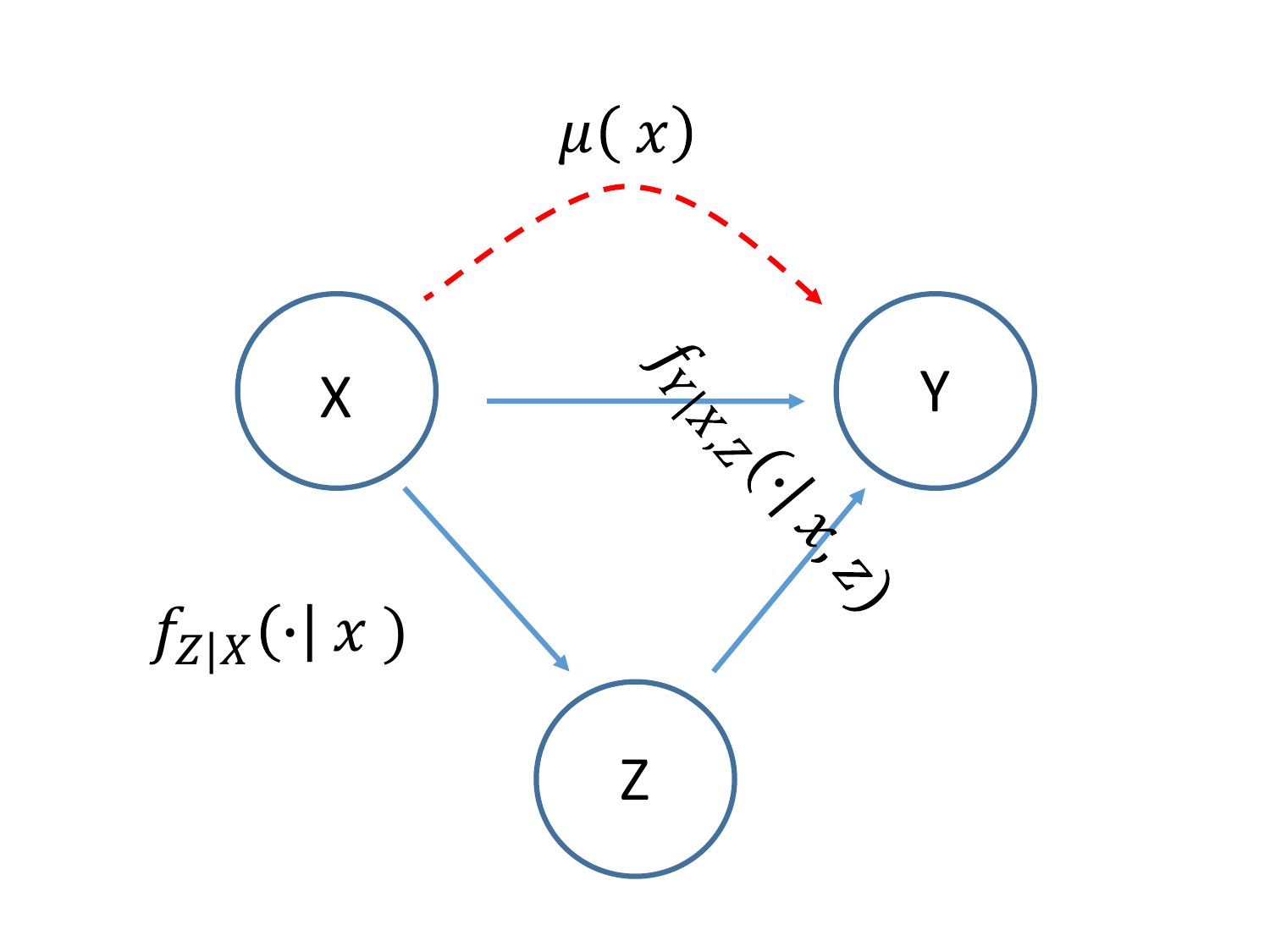}
    \caption{Errors in $Y$ setting} % war: Errors in Variables setting
    \label{fig:errory}
\end{figure}

Errors may occur not only in the features but also in the outcome. 
We focus here on a binary outcome, the typical framework in classification. (See Supplementary Material \cite{supp} on 
why regression, i.e., quantitative outcome variables, can also be affected, but potentially less so.)
%We briefly comment on regression, i.e., quantitative outcome variables in Appendix~\ref{subsec:app.errors.in.y.comments.on.regression}.) 
Among the multiple causes for errors in $Y$, labelling, in particular human annotation stands out. (Of course, some features $X$ may also require labelling. Yet, it is plausible that the fact that a variable cannot be measured easily, e.g., with a technical device, is the very reason why one is interested in a prediction model for $Y$ in the first place. Hence, labelling has a stronger connection to outcome variables than to input variables.) 
In tasks in which there exists a single, objective true value $y$, one reason for ``noisy labels'' is imperfect intra- and inter-annotator reliability \citep{frenay.verleysen.2014.label.noise}. In addition, the design of annotation instruments (e.g., instructions to the annotators) can affect the bias and variation of labels \citep{kern-etal-2023-annotation.acl, beck-etal-2024-order.acl}.

Numerous applications want to predict the true value of the variable of interest instead of an error-prone version of it. 
%of said variable: e.g., we want to predict a person's \emph{true} income, and not the level of income that this person would \emph{report} in a data collection analogous to that of our training data.
We show subsequently that model predictions are biased if we do not measure the outcome variable without error, except when the different error components happen to cancel each other out exactly for all values. This condition is very unlikely to hold for categorical outcomes as even random, unsystematic errors in measurement, labelling, or processing produce this bias.

Let $Z \in\{0,1\}$ be the true value of the outcome variable, i.e., without any error, and let $Y \in\{0,1\}$ now be the error-prone, but observed variable. $Y$ is not necessarily biased for $Z$; it is just not measured error-free. 
We would ideally want to learn $f_{Z|X}$, but have to settle for $f_{Y|X}$ as $Z$ is unobserved. Thus, the difference $f_{Y|X} - f_{Z|X}$ is the bias that we incur by having to train with $Y$ instead of $Z$.
We sketch the setup in Figure~\ref{fig:errory}.

Adapting \eqref{eq:datpred} to discrete $Z$ leads to
\begin{align}
	\Prob_{Y|X}(y|x) = \sum_{z \in {\cal Z}}  \Prob_{Y|X,Z} (y| x,z) \Prob_{Z|X}(z|x)  
	\label{eq:true_y_given_x_discrete}
\end{align}
where $\Prob(\cdot)$ denotes the corresponding probabilities. 
%Table~
\autoref{tab:true.false.Z.Y} summarizes our scenario: $Y$ and $Z$ can be in line, with true 1's or true 0's. There can also be two types of error: false 0's, with probability $\Prob_{Y,Z|X} (Y=0, Z = 1| X=x)$ conditional on $x$; and false 1's, with probability $\Prob_{Y,Z|X} (Y=1, Z = 0| X=x)$ conditional on $x$.
%(also called missed 1's)
\begin{table}[!bt]
	%\centering
	\caption{Relation of true values $Z$ and observed values $Y$ in binary classification.}
	\label{tab:true.false.Z.Y}
	\begin{tabular}{ cc|c|c| }
		& \multicolumn{1}{c}{} & \multicolumn{2}{c}{$Y$ (Observed)} \\
		& \multicolumn{1}{c}{}
		&  \multicolumn{1}{c}{1}
		& \multicolumn{1}{c}{0} \\
		\cline{3-4}
		\multirow{2}{*}{$Z$ (Truth)}  & 1 & true 1  & false 0  \\
		%	  &   &    	  & \red{missed 1 (missed 1)}) \\
		\cline{3-4}
		& 0 & false 1 & true 0	  \\
		\cline{3-4}
	\end{tabular}
\end{table}

As we show in Appendix~\ref{subsec:app.errors.in.y.bias}
\begin{align}
\label{eq:true.false.Z.Y.bias}
	\Prob_{Y|X}&(Y=1|X=x)  = \\ 
    & \Prob_{Z|X}(Z=1|X=x) \nonumber  \\ 
    & +  \Prob_{Y,Z|X} (\underbrace{Y=1, Z=0}_{\text{false $Y=1$}}| X=x) \nonumber \\ 
    & - \Prob_{Y,Z|X} (\underbrace{Y=0, Z=1}_{\text{false $Y=0$}}| X=x). \nonumber
\end{align}	
That is, a model trained with data $(Y,X)$ instead of $(Z,X)$ learns not just the desired $\Prob_{Z|X}$, but also two additional terms: the probability of false 1's (conditional on $x$), which on their own produce a positive bias, and the probability of false 0's (conditional on $x$), which produce a negative bias. The difference between these two error probabilities is the bias incurred by training with $Y$ instead of $Z$. Thus, training with $Y$ is only unbiased for predicting $Z$ when the two errors are exactly equally likely -- for every value $x$ -- which translates into a very strong set of assumptions. \eqref{eq:true.false.Z.Y.bias} generalizes to multi-class prediction (Appendix~\ref{subsec:app.errors.in.y.bias}), but then the number of assumptions required for unbiasedness even multiplies.

Note that the two error probabilities in \eqref{eq:true.false.Z.Y.bias} are joint probabilities of $Y$ and $Z$, $\Prob_{Y,Z|X}$, and not probabilities that are conditional on $Z$, which would be denoted $\Prob_{Y|X,Z}$. 
%This is not just a technical detail. Many real-world error sources actually fit specifically the conditional on $Z$ perspective: for instance, we might assume clerical error rates in manual data entry to be {the same for each value of $Z$ (and of $X$)} when humans randomly generate typos, enter data in the adjacent column or row instead of the correct cell, or click on the adjacent button, regardless of $Z$.\footnote{Similarly, we would expect the probability of random errors from annotators, respondents, or interviewers, especially when tired, inattentive, or low on motivation, to be the same for each value of $Z$ and of $X$.}
Joint and conditional error probability are however linked: $\Prob_{Y,Z|X} = \Prob_{Y|X,Z} \cdot \Prob_{Z|X}$. This shows (see Appendix~\ref{subsec:app.errors.in.y.cond.error}) that even if we assume that the conditional error probabilities are identical, as in, e.g., human-generated random typos or labelling errors, i.e., 
\begin{align}
 \label{eq:true.false.Z.Y.bias.same.cond.error}
	&\Prob_{Y|X,Z} (\underbrace{Y=1|X=x, Z=0}_{\text{false $Y=1$}}) =& \\
	&\Prob_{Y|X,Z} (\underbrace{Y=0|X=x, Z=1}_{\text{false $Y=0$}})  \eqqcolon c(x),& \nonumber 
\end{align}
the bias does not vanish unless the marginal conditional probabilities  
	$\Prob_{Z|X}(Z=0|X=x)$ and $\Prob_{Z|X}(Z=1|X=x)$ 
are also identical.  With equal and positive conditional error probability $c(x)>0$, unbiasedness is therefore only possible in binary classification iff
\begin{align}
	\Prob_{Z|X}(Z=1|X=x) = 0.5 = \Prob_{Z|X}(Z=0|X=x) ,
	\label{eq:true.false.Z.Y.bias.conditional.all.0.5}
\end{align}
i.e., the binary classification problem must be exactly balanced for all $x$ and $Z$ must be independent of $X$, rendering the whole exercise of data analysis moot. In short, equality of conditional error probabilities might be common in practice but is rather harmful than helpful in terms of bias. 
Conversely: for fixed, equal conditional error probabilities $c(x)>0$, the less balanced the true classes, the greater the resulting bias \citep[see Supplementary Material][]{supp}. 

Note also that the discussed bias is a form of model uncertainty that cannot be reduced by increasing sample size. All three components in \eqref{eq:true.false.Z.Y.bias}, including the two bias components, are also subject to (reducible) random sampling uncertainty.
\subsection{The Non-\textit{i.i.d.} Setup}
\label{subsec:non.iid.data}

Standard machine learning as well as statistical models build upon the standard setup that available data are drawn  \textit{i.i.d.} from (\ref{eq:datagen}), leading to the data source ${\cal D}=\{ (y^{(i)} , x^{(i)}), i= 1, \ldots , n\}$. This assumption is often violated and, if so, one needs to carefully adjust the training algorithm for that. For instance, when data are collected over time and the aleatoric model (\ref{eq:aleatoric}) changes over time, the 
\textit{i.i.d.} assumption does not hold. 
We will look deeper into such changes in Section~\ref{subsec:deployment}. 
Here we focus on constellations where data are clustered and the assumption of independence is violated.
As an illustrative example, we consider images $X$ for which human labellers or annotators provide the corresponding labels $Y$. 
 
Let us first construct an \textit{i.i.d.} setup for this example. We consider each image $X$ to be randomly drawn from a set of infinitely many images. Each image $X$ is labelled by a different randomly chosen labeller.
Some crowd-sourcing initiatives in which images are classified by lay persons are approximately described by this setup
\citep{northcutt2021confident}. Possible uncertainties in the labelling are often incorporated as so-called label
noise \citep{rolnick2017deep}. For a general discussion of labeller uncertainty see also \citet{geva.et.al.2019.modeling.task.or.annotator} or \citet{misra.et.al.2016.noisy.human.labels.in.visual.data}.

However, image classification may require expertise or training: e.g., when medical images are assessed with respect to some disease (see, e.g., \citealp{zhou2021review} and \citealp{dgani2018training}). Ambiguity occurs if different experts come to different conclusions and to account for this, one image is sometimes labelled by multiple labellers.
The latter violates the assumption of independence among labels. If, for now, we ignore any dependence due to the labellers, we can accommodate this data constellation by extending the framework from above. If each image $x^{(i)}$ is labelled by $J$ labellers, we may denote the labels as $y^{(i)}_1, \dots , y^{(i)}_J$ leading to  the  multivariate outcome $y^{(i)} = (y^{(i)}_1, \dots , y^{(i)}_J)$. Hence, in principle we are back in the \textit{i.i.d.} setup
\eqref{eq:datagen} but with a
multivariate response variable. Letting $y_j^{(i)} \in \{ 1, \ldots , K\}$ for all $j = 1, \ldots ,J$ suggests to model  $f_{Y|X}(y|x)$ as multinomial or a Dirichlet-multinomial distribution, which
requires that the entire label vector $y^{(i)} = (y^{(i)}_1, \ldots , y^{(i)}_J)$  is used for training. Often, however, this multivariate framework is ignored and the outcome is reduced to a scalar, i.e., the majority class among $(y^{(i)}_1, \ldots , y^{(i)}_J)$,
see, e.g., \citet{zhu2020so2sat} or \citet{rodriguez2018beyond}. 
It is important to note that such an approach ignores information about the level of uncertainty expressed in different labels given by the labellers. 
Recent work confirms that processing all votes takes uncertainty more appropriately into account, see, e.g., \citet{peterson2019human} or \citet{battleday2020capturing}.

The above setup refers to a single observation, i.e., each image is labelled by the same $J$ labellers. Typically, however, not every annotator labels all the images. That is, the data are not only clustered by images, with each image being labelled by multiple labellers, but often also clustered by labellers, i.e., each labeller labels multiple images. In statistics, this is typically known as repeated measurements  
\citep{davis2002statistical} or multilevel, hierarchical, nested, or mixed models \citep{goldstein2011multilevel}.
Labellers can differ in their level of uncertainty or reliability (variance) and their central tendency (bias), which carries forward to the data. In other words, the heterogeneity of labellers violates the \textit{i.i.d.} assumption. 
If such labeller effects were included in the model, aleatoric uncertainty could be reduced (see~\ref{subsubsec:omitted.variables}). In addition, labellers may not be randomly assigned to units: e.g., lay annotators may self-select which images they choose to label. If such labeller effects are unaccounted for, the trained model may be biased (similar to unit nonresponse bias, see Section~\ref{subsec:missing.data}). 
Moreover, in most machine learning models the predictions $\hat{p}(y|x_0)$ are much more influenced by data points that are close to $x_0$ than by those further away. Thus, to get at the actual level of local epistemic uncertainty we must focus more on the uncertainty of labels of observations close to $x_0$ in the training data instead of relying on global label uncertainty measures.
However, there is also epistemic uncertainty in estimating labeller effects. Hence, a sufficiently large number of labelled images from each labeller may be required.

In sum, non-\textit{i.i.d.} scenarios can be complex and the described labelling problem just serves as an example of how quickly independence, as well as identical distributions, are violated in the data used for training machine learning models. In general, this is still an open field for future research.

% Kapitel 5
\section{Data Uncertainty}
\label{sec:data-uncertainty}
 
\subsection{General Comments}\label{subsec:data.uncertainty.general.comments}

Data are omnipresent in many machine learning applications, and high data quality is key. Yet, following the literature, we have discussed aleatoric and epistemic uncertainty in terms of probability distributions, building on the idealized model that data are generated as identical draws from some probability distribution. To properly deploy machine learning models in real-world settings, data scientists need some knowledge about the data production process itself. Are the data suitable for the purpose intended by the analyst?

This judgement call is often difficult to make as data production can be a rather complex endeavour. One can hardly control all aspects of it. 
Data collection protocols are only specific up to a point and may lack relevant detail. The social context as well as the technologies in use may influence the data production process and change over time. Data can have all sorts of deficiencies, failing to provide the desired virtual copy of the real world. The uncertainty about how a given dataset relates to the real world, including the unknown factors of what happened during its creation, is what we call ``data uncertainty''. This all-embracing definition is meant to widen our perspective about the various kinds of uncertainties that exist, besides the ones discussed in machine learning.

The next subsection illustrates common challenges around data production, triggering data uncertainty. Remedies in the face of data uncertainty are only possible to some extent, as we exemplify afterwards for missing data and for data with shifting distributions.
\subsection{Total Survey Error}
\label{subsec:tse}

When faced with data it is easy to overlook the fact that data are always designed, deliberately or unintentionally, although the eventual data analyst might be uninvolved in or even unaware of decisions going into the data generating process  (\citealp{groves.2011.blog.designed.organic.data,groves.2011.3.eras.of.survey.research}, 
\citealp[p.~843]{japec.kreuter.biemer.lane.et.al.2015.big.data.survey.research.aapor.task.force.report}). 
	
\begin{figure*}[!tb]
\begin{minipage}[t]{0.45\linewidth}

%	\centering
        \vspace{0pt}    
	\includegraphics[width=1\linewidth]{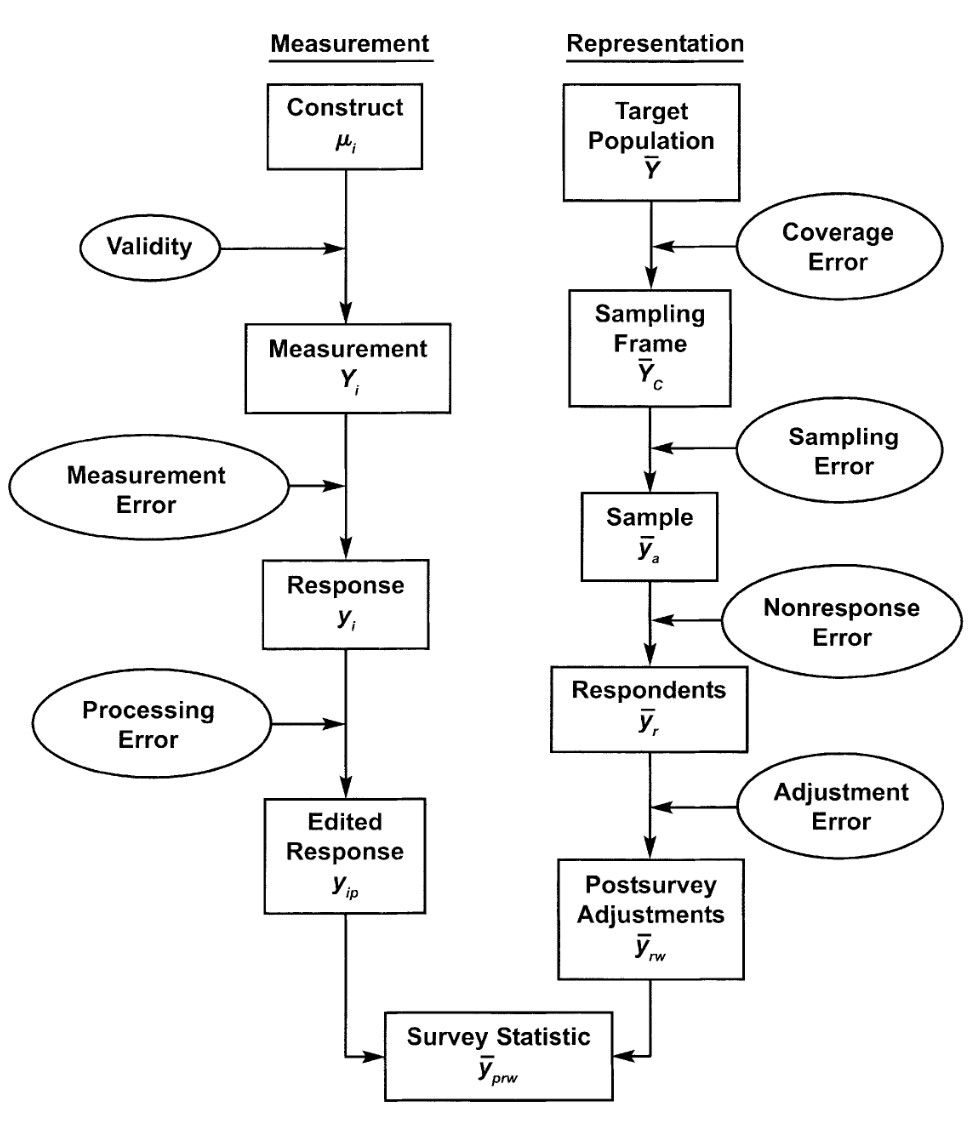}
 \end{minipage}
 \begin{minipage}[t]{0.45\linewidth}
  \vspace{0pt}
\includegraphics[width=1\linewidth]{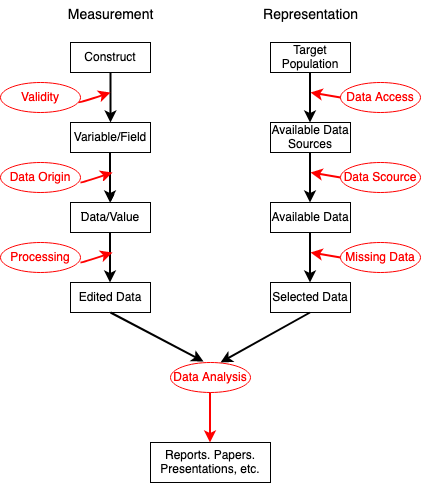}
 \end{minipage}
	\caption{Left: Total Survey Error Framework Components \citep[copied from][]{groves_survey_2009}, Right: Total Data Quality Framework in Machine Learning %\citep[copied from][CC BY-NC 4.0]{west.wagner.2023.tdq}
 (Dimensions of TDQ by \citealp{west.wagner.2023.tdq} is licensed under \href{https://creativecommons.org/licenses/by-nc/4.0/}{CC BY-NC 4.0}).}
	\label{fig:Total_Survey_Error}
\end{figure*}

For the special case of survey data, survey methodologists and survey statisticians have used the Total Survey Error (TSE) framework (see Figure~\ref{fig:Total_Survey_Error}, left-hand side) to stay attuned to decisions going into the production of survey statistics and errors arising alongside \citep{andersen_total_1979, groves_survey_2009, groves_total_2010}. Such parsing of data generating processes can help for all data types (tabular, semi-, and unstructured) to identify aleatoric and epistemic uncertainties. The framework is also employed to improve the design and the collection of data. Aside from active learning, considerations to address data uncertainties at the design and collection stage are mostly absent from the more theoretical literatures in statistics and machine learning. 
In the design phase, what is aleatoric because of, e.g., omitted variables (see Section~\ref{subsubsec:omitted.variables}) and what is not, is not yet set in stone.

\phantomsection\label{anchor.tse.representation.and.measurement.sides}

\citet{west.wagner.2023.tdq} extended the framework to be generally applicable beyond survey data in their Massive Open Online Course (see Figure~\ref{fig:Total_Survey_Error}, right-hand side) and highlight examples of the various error sources in typical machine learning data. 
In this framework, \emph{error} denotes deviations of what we have (e.g., data) from the truth. Or, in circumstances in which ``truth'' is not an adequate concept: deviations from what we intended to capture. Thus, \emph{error} includes systematic differences, i.e., bias, but also non-systematic variation.

The widely used TSE framework and its extensions summarize the various possible sources of error that can arise during data production. It invites researchers to think through and compare the errors' relative relevance in their own data. It makes clear that datasets are often not the perfect abstraction of the real world that researchers would like to have: 
we have data uncertainty. This can be highly relevant in practical machine learning applications, though hard to take hold of in statistical and machine learning models. 
\subsection{Missing Data}   \label{subsec:missing.data}

Data are often not complete, meaning that data entries are missing. We therefore dedicate this section to illuminating the role of missing data in the context of machine learning. Generally, one distinguishes between two types of missing data. First, \emph{unit nonresponse} is when no data are collected at all for a unit, that is the entire data entry $(y^{(i)},x^{(i)})$ is not available for unit $i$. 
Second, if some but not all values in 
$(y^{(i)},x^{(i)})$ are missing, this is called \emph{item nonresponse}: 
i.e., some cells for the $i$-th data entry are empty. For instance, certain parts of a text $X$ or some pixels of an image $X$ are missing, or an annotator did not provide a label $Y$ for a specific unit. Unit nonresponse is usually linked to the representation side in the TSE framework (Section~\ref{anchor.tse.representation.and.measurement.sides}), whereas item nonresponse is mostly related to the measurement side. 
Moreover, record linkage of two disparate data sources can cause unit nonresponse (when keeping only units that are recognized to be in both data sources) or item missingness (when keeping all units, those who are only in one data source exhibit item nonresponse on the variables from the other source).

The missing data problem is important for several reasons. The main worry is that units that suffer from nonresponse tend to be different from those that do not: as we address below, this can affect uncertainty even beyond the well-known selection bias. 
Second, there is also a practical mandate: unless the missing values are filled in, no prediction can be made for units with item nonresponse in features that are part of the prediction model. 
Imputation models \citep[Ch.~4f.]{little2019statistical} estimate the missing value based on the observed information, e.g., via regression of one feature on all the other features. 
A third reason is efficiency. By default, most standard software packages ignore all data from units with missing entries, which is called naive complete-case analysis. 
This is typically not efficient, because not all available information is used, and epistemic uncertainty is elevated \citep[Ch.~3.2]{little2019statistical}. In a sense, this problem is more pronounced in settings for machine learning than for traditional statistics \citep[see Supplementary Material][]{supp}.

Let $R^{(i)}$ be a response indicator random variable which equals $1$ if all components of $\left( y^{(i)},x^{(i)} \right)$ are observed and $0$ otherwise, i.e., when at least one component or the whole unit $i$ is missing. Note that the complete-case data consist of $(y^{(i)},x^{(i)},R^{(i)}=1)$ so that any prediction model trained with naive complete-case analysis will learn $f_{Y|X,R}(y|x,R=1)$ instead of the population average%$f_{Y|X}(y|x)$.
\begin{align}
        f_{Y|X}(y|x) = \sum_{r \in \lbrace 0,1 \rbrace}f_{Y|X,R}(y|x, R=r) \Prob(R=r|x).
    \label{eq:missing.population.model}
\end{align}
Because of its focus on population inference, traditional statistical analysis targets $Y|X$, which we call the ``population model''. Predictive modelling is interested in learning $Y|X$ when the goal is to predict and predict well for all units, including those affected by missingness. 
We thus compare the population model $Y|X$ and the model from naive complete-case analysis $Y|X,R=1$. Unless noted otherwise, mathematical derivations 
are analogous to~\ref{subsubsec:biases.due.to.unobserved.variables.general.framework} and~\ref{subsubsec:omitted.variables} 
with $R$ in the role of $Z$. 
In Appendix~\ref{sec:app.proof.of.eq.missing.data.mechanism} we show that
\begin{align}
    f_{Y|X,R}(y|x,R=1) = \underbrace{\frac{\Prob (R=1 | y , x )}{\Prob(R=1 | x )}}_{\mbox{bias factor}} f_{Y|X}(y|x).
    \label{eq:missing.data.mechanism}
 \end{align}
Naive complete-case analysis yields the population model iff the bias factor equals 1, i.e., when the missingness $R$ is conditionally independent of $Y$ given $X$. This is true for item missingness in $Y$, or in $X$, or in both, or for unit nonresponse. %This implies that if the missingness is in the response variable $Y$ only and $R$ is conditionally independent of $Y$ given $X=x$, the bias factor equals $1 $ and a complete case analysis targets the population model. 
 
A more general discussion on missing data is found in the 
 pertinent statistical literature \citep[Ch.~1.3]{little2019statistical}, where missingness is categorized into three patterns. The first, \emph{missing completely at random} (MCAR) defines the setting in which missingness depends on neither $Y$ nor $X$, that is 
 \[
 \Prob (R=1 | y , x ) = \Prob(R=1) = \Prob(R=1 | x ).
 \]
This is sufficient, but not necessary for unbiased naive complete-case analysis, as stated above. Second,  data are called \emph{missing at random} (MAR), if the propensity of missingness depends only on the observed but not on the missing values. MAR is a helpful concept for univariate analysis of $Y$, but not with regard to $Y|X$: MAR is neither necessary nor sufficient for unbiasedness of naive complete-case analysis -- but the above-mentioned independence 
$R \!\perp\!\!\!\perp Y | X$ is. 
 %of missingness and $Y$ given $X$ is. 
 Also, in machine learning, even imputation is problematic when $R \not\!\perp\!\!\!\perp Y | X$, see Appendix~\ref{subsec:app.discussion.of.mar}. 
 Finally, the third missingness pattern is \emph{missing not at random} (MNAR), when the propensity of missingness (also) depends on the missing values. This is generally a problematic scenario that implies an unspecifiable bias.

Let us now look at the  aleatoric uncertainty in the population model expressed through the variance
\begin{align}
\label{eq:missing.aleatoric.variance}
     Var_{Y|X}&(Y|x)  =  E_{R|X}\left( Var_{Y|X,R}(Y| x, r)\right)  \\
     &+  E_{R|X}\left( bias(x,r)^2 \right)  \nonumber \\
              =&  \sum_{r \in \lbrace 0,1 \rbrace} \Prob(R=r|x) Var_{Y|X,R}(Y| x, r) \nonumber \\
             &+ \sum_{r \in \lbrace 0,1 \rbrace} \Prob(R=r|x)  bias(x,r)^2, \nonumber
\end{align}
where, for $r \in \lbrace 0, 1 \rbrace$, 
\begin{align}
\label{eq:bias.between.respondents.and.population.average}
bias(x,r)  \coloneqq & E_{Y|X,R}(Y|x,R = r) - E_{Y|X}(Y|x)  \\
           = & \left[ 1 - \Prob\left(R=r|x\right)  \right] \bigl[ E_{Y|X,R}(Y|x, R = r) \nonumber \\
           &- E_{Y|X,R}(Y|x, R \neq r) \bigr]. \nonumber
\end{align}
First, instead of the conditional variances of the respondents and nonrespondents respectively, the population model uses their weighted average. Investigating such variance heterogeneity, possibly aided by auxiliary variables $W$ (see Appendix~\ref{missing.data.auxiliary.data.w}), could improve pointwise predictive uncertainty quantification. Label noise being correlated with response propensity is one possible cause.\phantomsection\label{interaction.nonresponse.measurement.error} A more holistic approach such as the TSE framework (\ref{subsec:tse}) makes it easier to recognize such possible ``interactions'' among seemingly unconnected sources of uncertainty -- here, nonresponse and measurement error. 

Second, the population model exhibits bias for both, respondents and nonrespondents, if the respective conditional expectations $E_{Y|X,R}$ differ, which is typically caused by $f_{Y|X,R}(y|x, R= 1) \neq f_{Y|X,R}(y|x, R= 0)$, 
%\[
%f_{Y|X,R}(y|x, R= 1) \allowbreak \neq f_{Y|X,R}(y|x, R= 0).
%\] 
which by \eqref{eq:missing.data.mechanism} is equivalent to $R \not\!\perp\!\!\!\perp Y | X$ (see Appendix~\ref{sec:app.proof.of.eq.missing.data.mechanism}). 
Setting variance heterogeneity aside, this bias is why aleatoric uncertainty is generally larger in the population model. However, in addition to the aforementioned efficiency, one rationale for the population model is that it is more suitable for nonrespondents than the naive complete-case model would be, regarding both, variance and expected value. 

Finally, as in previous sections, we see that what is considered aleatoric uncertainty depends on context: here, on how we choose to handle missing data -- and the choices afforded to us by the data situation. 
In practice, $f_{Y|X}$ may need to be recovered via weighting or imputation models. To do so, assumptions about the missingness mechanism are required, as the true mechanisms often remain unknown (data uncertainty). The aleatoric and epistemic uncertainty affecting the weighting and imputation models are considered in the Supplementary Material \cite{supp}.
\subsection{Deployment}\label{subsec:deployment}
\subsubsection{General Setup}

When machine learning models are deployed in real-world applications, possible shifts in the data need to be dealt with. For example, new measurement protocols may be in place, or in a new data source the relations learned by the model may no longer be valid, for instance, due to changes in true relationships or due to changes in the data deficiencies and error mechanisms discussed in Section~\ref{subsec:tse}.
    \phantomsection\label{change.in.error.mechanism}
It may also occur that
 the model predictions lead to actual change in real-world behaviours \citep{perdomo.et.al.2022.performative.prediction}. 
Looking just at the conditional distribution $f_{Y|X}(y|x)$ may, thus, oversimplify many issues in practical machine learning applications,
as the implicit assumption that all observations are drawn independently from one single distribution (\textit{i.i.d.}) does not always hold. More realistically, we have one training dataset $\mathcal{D}^{{\cal T}r} = \{(x^{(i)}, y^{(i)}), i=1, \ldots, n_{Tr}\}$, which is used to train or fit the model and a separate deployment dataset $\mathcal{D}^{{\cal D}e} = \{(x^{(i)}, y^{(i)}), i=1, \ldots, n_{De}\}$, on which the previously trained model will be deployed. Note that we want to make a clear distinction between testing and deployment. 
While $Y$ is readily available in the training data, it is typically unavailable during deployment at the time when $Y$ is predicted. Before deployment, it is common to assess the quality of the model and its predictions. For this purpose, the trained model is evaluated on test or validation data for which $Y$ is available, often obtained by a random train-test split of $\mathcal{D}^{{\cal T}r}$.
Both training and deployment data can come from different sources and do not need to be representative of the same population.
For this reason, our notation emphasizes that the joint distribution can be different in $\mathcal{D}^{{\cal T}r}$ and $\mathcal{D}^{{\cal D}e}$.

We want to formalize the previously described problem. Training observations are drawn from a training (super) population according to 
\begin{align}
    \label{eq:true_train_xyz}
      (X,Y,Z)^{{\cal T}r} & \sim  f_{X,Y,Z}^{{\cal T}r}(x,y,z) \\
      & =  f_{Y| X,Z}^{{\cal T}r}(y|x,z) \: f_{X|Z}^{{\cal T}r}(x|z) \: f_Z^{{\cal T}r}(z) \nonumber    
\end{align}
and the deployment observations are drawn from 
\begin{align}
    \label{eq:true_deploy_xyz}
      (X,Y,Z)^{{\cal D}e} & \sim  f_{X,Y,Z}^{{\cal D}e}(x,y,z) \\
      & =  f_{Y| X,Z}^{{\cal D}e}(y|x,z) \: f_{X|Z}^{{\cal D}e}(x|z) \: f_Z^{{\cal D}e}(z). \nonumber    
\end{align}
The joint distributions $f^{{\cal T}r}_{Y, X}$ and $f^{{\cal D}e}_{Y, X}$ follow from \eqref{eq:true_train_xyz} and \eqref{eq:true_deploy_xyz} analogously to the derivation of $f_{Y, X}$ in \eqref{eq:datagen}. 
In the same manner, the conditional distributions are
\begin{align}
f^{{\cal T}r}_{Y|X}(y|x) &=&
\int_{z \in {\cal Z}} f_{Y|X,Z}^{{\cal T}r}(y|x,z) \: f_{Z|X}^{{\cal T}r}(z|x) dz
\label{eq:datpred_train}
\end{align}
and 
\begin{align}
f^{{\cal D}e}_{Y|X}(y|x) &=& \int_{z \in {\cal Z}} f_{Y|X,Z}^{{\cal D}e}(y|x,z) \: f_{Z|X}^{{\cal D}e}(z|x) dz.
\label{eq:datpred_deploy}
\end{align}
\phantomsection\label{text.prediction.model}
A machine learning model is trained with data from $\mathcal{D}^{{\cal T}r}$ to approximate $f^{{\cal T}r}_{Y|X}$; we denote the trained \textit{prediction model} as $\hat{p}^{{\cal T}r}(y|x) = p(y|x ; \hat{\theta}^{{\cal T}r})$.

In the following section, we study the conditions under which $f^{{\cal T}r}_{Y|X}$ and $f^{{\cal D}e}_{Y|X}$ are identical or possibly different, a common challenge in empirical work.

\subsubsection{External Validity, Transportability, and Distribution Shift}          
\label{subsec:distribution.shift}

The available data ($\mathcal{D}^{{\cal T}r}$) are only loosely related to the environment we are interested in ($\mathcal{D}^{{\cal D}e}$). There is thus uncertainty about how well the causes and correlations found in data from the first environment translate to related situations, i.e., (how well) are the findings from $\mathcal{D}^{{\cal T}r}$ replicable in a different environment?

Since experiments take place in highly controlled settings, practitioners often wonder if the lessons learned therein would still hold true beyond the experimental context. \citet{campbell_experimental_1963} take up this concern and examine the validity of various experimental designs, drawing attention to \textit{external validity}, i.e., factors impeding the generalizability of an experiment. 
One would wish to take a model, including all its estimated parameters, from the first environment and apply it in a second environment without bias. This feature of a model has been called \textit{transportability} (e.g., \citealp{pearl_external_2014} in the context of structural causal models and \citealp[Ch.~2.2.4f.]{carroll.et.al.2006.measurement.error} in the measurement error literature). 
It requires that the same relations between variables are present in each of the environments. If the environments differ in uncontrolled ways, for example, because observational units are allowed to enter a study population through unknown processes or due to any other source of error (see the TSE framework in Section~\ref{subsec:tse}), external validity and transportability are at risk \citep{keiding_perils_2016, keiding_web-based_2018, egami.hartman.2022.external.validity}.% 

In the machine learning literature, \textit{distribution shift}, \textit{dataset shift}, or \textit{concept drift} occurs if the joint distribution of inputs and outputs differs between the training and deployment stage, i.e., $f^{{\cal T}r}_{Y, X} \neq f^{{\cal D}e}_{Y, X}$ (e.g., \citealp{quinonero-candela2009DatasetShiftMachine, gama_survey_2014, varshney2022TrustworthyMachineLearning}). Note that the terminology and the exact definitions vary. 
Sometimes there is no differentiation between \textit{testing} and \textit{deployment} and distribution/dataset shift is defined as differences in the joint distribution of $Y$ and $X$ between training and \textit{testing}. We are, however, interested in the actual application/deployment.
As a consequence, the conditional distributions $f^{{\cal T}r}_{Y| X}$ and $f^{{\cal D}e}_{Y| X}$ will often differ as well.

This lack of transportability is a critical problem for predictive models, as what the model learned during training might not be valid during deployment any more and the trained model will need to be updated.
We discuss several scenarios and how they lead to transportability next, especially with regard to the role of $Z$. We, therefore, assess whether the identity $f^{{\cal T}r}_{Y|X} = f^{{\cal D}e}_{Y|X}$ holds, i.e., when is the conditional distribution $f^{{\cal T}r}_{Y|X}$  estimated from the training data identical to the one present during deployment, $f^{{\cal D}e}_{Y|X}$?

\subsubsection{Identical Super Population}

First, if the two samples $\mathcal{D}^{{\cal T}r}$ and $\mathcal{D}^{{\cal D}e}$ are created in identical ways, it follows that $f_{X,Y,Z}^{{\cal T}r} = f_{X,Y,Z}^{{\cal D}e}$ and,
then, trivially, the respective conditional distributions are identical. 
One simple example of $f_{X,Y,Z}^{{\cal T}r} = f_{X,Y,Z}^{{\cal D}e}$ is when an initial dataset is randomly split into multiple parts, some of which are used for training and others for testing. This train-test split does not correspond to what is meant by deploying a model but oftentimes ensures identical distributions. This is the scenario that is underlying standard cross-validation. 

\subsubsection{Component-wise Equivalence}
Second, if both components on the right-hand side of \eqref{eq:datpred_train} and \eqref{eq:datpred_deploy} are equal, i.e., $f^{{\cal T}r}_{Y|X,Z} = f_{Y|X,Z}^{{\cal D}e}$ and $f^{{\cal T}r}_{Z|X} = f_{Z|X}^{{\cal D}e}$, then $f^{{\cal T}r}_{Y|X} = f^{{\cal D}e}_{Y|X} $, too. This leads directly to transportability. Conversely, if at least one of the components differs between \eqref{eq:datpred_train} and \eqref{eq:datpred_deploy}, then $f^{{\cal T}r}_{Y|X}$ and $f^{{\cal D}e}_{Y|X}$ will usually differ. Note that equal marginal distributions of $X$, $f^{{\cal T}r}_X = f^{{\cal D}e}_X$, are not required.

\subsubsection{Independence of $Z$}
Third, if in both populations $Y$ is independent of $Z$ given $X$, denoted as $Y \perp \!\!\! \perp Z | X$, i.e., $f^{{\cal T}r}_{Y|X,Z} = f^{{\cal T}r}_{Y|X}$ and $f^{{\cal D}e}_{Y|X,Z} = f^{{\cal D}e}_{Y|X}$, then the right-hand sides in (\ref{eq:datpred_train}) and (\ref{eq:datpred_deploy}) simplify to $f^{{\cal T}r}_{Y|X}$ and $f^{{\cal D}e}_{Y|X}$, respectively. This implies that the distributions of $Z$, $f_{Z|X}^{{\cal T}r}$ and $f_{Z|X}^{{\cal D}e}$, become irrelevant. In other words: to learn the conditional distributions $f^{{\cal T}r}_{Y|X}$ and $f^{{\cal D}e}_{Y|X}$, variables $Z$ which contain no additional information about $Y$ beyond what is already contained in $X$ can be safely ignored -- even when $X$ is not independent of $Z$.
This might be the case when $Y$ is obtained by annotating $X$, e.g., in image classification, where human annotators do not have access to $Z$, and the labels $Y$ rely entirely on the images $X$.
Yet, assuming $Y \perp \!\!\! \perp Z | X$ does not yield transportability on its own, as the simplified distributions $f^{{\cal T}r}_{Y|X}$ and $f^{{\cal D}e}_{Y|X}$ can still differ.

As a recipe to ensure $f^{{\cal T}r}_{Y|X,Z} = f^{{\cal D}e}_{Y|X,Z}$ holds, one may include in the vector $(X, Z)$ all variables that could possibly affect $Y$ (the unobserved $Z$ is likely to be high-dimensional). Such a model would make, apart from estimation uncertainty, an optimal prediction $\hat{p}^{{\cal T}r}(y|x, z)$ for every single observation, meaning that the predictive accuracy of $f_{Y|X,Z}$ could not be improved upon. In particular, as long as $f^{{\cal T}r}_{Y|X,Z} \neq f^{{\cal D}e}_{Y|X,Z}$, there are additional environment-related variables $Z$ that affect $Y$ and that would need to be included, until 
\[
f_{Y|X,Z} = f^{{\cal T}r}_{Y|X,Z} = f^{{\cal D}e}_{Y|X,Z}.
\]
One might call $f_{Y|X,Z}$ the true model since the relation remains the same across environments. In practice, however, this model can hardly be built, since the constructs that would need to be included in $(X, Z)$ are rarely known, and their measurement is an even greater challenge. Extensive expert knowledge would be needed to develop this model, even if just a very rough approximation of it may be feasible. In fact, the recipe provided has little practical value, but its argument should remind readers that detailed (expert) knowledge is key to finding all the relevant and measurable variables $X$ that ensure transportability. All kinds of uncertainty have to be dealt with, and so the model may even focus on the stylized facts while neglecting predictive performance. 

\subsubsection{Examples}
Consider image classification of animal photos as an example. Let therefore $X$ be the pixels in a photo of an animal, $Y$ the annotated label of the depicted animal, and $Z$ its true species.
We take a convenience sample of animals living in European cities, implying that some of the most frequent species in $f_Z^{{\cal T}r}$ are cats, dogs, and pigeons. In addition, the training data contains photos of elephants and giraffes taken in a zoo. Most annotators will easily identify the correct animal from the picture, but some unintentional labelling errors might occur. Also, some dogs may look like cats and confusion would increase if pictures of wolves and wildcats were included as well. The probabilistic framework allows for such mislabelling of species.
As we can only make use of the pictures to obtain labels (a zoological examination of all animals to determine the true species $z$ is not possible), it is safe to assume $f^{{\cal T}r}_{Y|X,Z} = f^{{\cal T}r}_{Y|X}$, at least if the annotators do not know the location from which the pictures originate.
Moreover, due to reasonable diligence among annotators and shared knowledge of what each animal looks like, we have no reason to believe that the labelling decisions in our training dataset are in any way special, but are representative of decisions others would make if they had images like ours.
Therefore, $f_{Y|X,Z}^{{\cal T}r} \approx f_{Y|X,Z}^{{\cal D}e}$ is a reasonable approximation.

Let us now consider the deployment of the model. A photographer decides to train a machine learning model with animal pictures taken in Europe. She wants to use the trained model $\hat{p}^{{\cal T}r}(y|x)$ to label new pictures from her latest safari tour. As we have argued, $f_{Y|X,Z}^{{\cal T}r} = f_{Y|X}^{{\cal T}r}$ and $f_{Y|X,Z}^{{\cal T}r} \approx f_{Y|X,Z}^{{\cal D}e}$ are reasonable assumptions. It thus follows that $f_{Y|X}^{{\cal T}r} \approx f_{Y|X}^{{\cal D}e}$. The photographer can therefore use the data from European cities to train a model for African animals from the safari and does, in theory with infinite-sized samples, not need to worry about distribution shift or transportability.

This framework depends on probability distributions, which would need to be estimated. In practice, there will be high estimation uncertainty (see Sections~\ref{sec:aleatoric_epistemic_uncertainty} and~\ref{sec:statistics_and_uncertainty}) for classes with little observed data and the model might perform better for animals that were more frequent in the training data. Transportability remains an issue.
Unequal subgroup accuracy might lead to varying uncertainty levels of a model and, thus, varying reliability. This would only cause minor inconvenience in an automated animal picture classification but can lead to major fairness issues, if, for example, humans are underrepresented according to sensitive attributes, like gender, age, race, or health status.

If however, the model is applied to animals that did not exist in the training data, e.g., images of zebras, we are facing issues with \textit{out of distribution} or \textit{out of data} (OOD) samples \citep{ren2019LikelihoodRatiosOutofDistribution, malinin2018PredictiveUncertaintyEstimation}. %\footnote{
Note that there are ambiguous definitions of the term out of distribution. See \citet{farquhar2022WhatOutofdistributionNot} for a discussion.%}

The two assumptions $Y \perp \!\!\! \perp Z | X$ and $f^{{\cal T}r}_{Y|X,Z} = f^{{\cal D}e}_{Y|X,Z}$ are often violated in practice. Consider, as another example, customer reviews about a product. The customers have their personal opinion $Z$ about the product, which remains an unobserved variable. The customer writes a review, the text data $X$. She also ticks a star rating $Y \in \{\mbox{1 star },\ldots , \mbox{5 stars} \}$ as a rough summary of how much she recommends the product to others. The question is: Would customers always describe their full reasoning for why they choose a specific star rating? Or would customers ``forget'' to mention some thoughts ($Z$) in the reviews, but still decide on a star rating based on unmentioned thoughts? If the latter is the case, $Y$ will depend not just on $X$ but also on $Z$. Therefore, the conditional independence assumption $f_{Y|X,Z}^{{\cal T}r} = f_{Y|X}^{{\cal T}r}$ would not hold in this example. Similar issues will arise whenever the labellers, i.e.,  the persons selecting the $Y$-values, have access to relevant information that is not included in $X$. For instance, a labeller annotates a low-resolution image $X$ and classifies this into category $Y$, but she has a high-resolution image $Z$ as an additional information source available. 

In our final example, occupation coding, scientists and statistical institutes around the world wish to measure and label the occupations people have. Simplifying the real challenges involved, $X$ is what respondents describe verbally when asked about their job in a survey. Their descriptions are often short, maybe just a job title, although, ideally, one would want to capture the ground truth $Z$ which is, roughly speaking, all the different tasks and duties people do in their job. 
Labelling experts select the most appropriate label $Y$ from a classification scheme, based on the input texts $X$. Since the labelling experts do not have access to the ground truth $Z$, the conditional independence assumption $f_{Y|X,Z}^{{\cal T}r} = f_{Y|X}^{{\cal T}r}$ holds.
However, social scientists skilled in content analysis are very much aware that accurate coding requires appropriate training. Not only are the labellers expected to follow formalized coding instructions that they learn during theoretical training, but they also adopt informal, unwritten rules from their peers during work \citep{hak_coder_1996}. 
\citet{massing_how_2019} report a 57\% agreement rate when experts from different institutes code the same occupations. Due to institute-specific factors, this rate increases substantially to $\sim$71\% and 84\%, depending on the institute, if agreement is measured between two experts from the same institute. All this means that a number of systematic factors related to the labelling process but unrelated to $X$ affect the final labelling decision. As the distributions of $f_{Y|X, Z}$ differ between institutes, the trained model is not transportable from one institute to the other. 
In addition, the low rates of agreement and the challenges of obtaining sufficiently large amounts of training data have hindered the large-scale deployment of machine learning models for occupation coding \citep{schierholz_machine_2020}.
Those examples showed how complex it might be to assess transportability in the real life deployment of predictive models, and thus the difficulty of assessing a model's uncertainty.

\section{Conclusion}
\label{sec:conclusion}
%\rone{Some practical methods, such as Hamiltonian Monte Carlo, which is considered as a golden standard method for posterior inference and frequently used for estimating parameters of Bayesian neural networks [6,8], and stochastic variational inference, which is commonly used in big data regime, have not been discussed, either.}
%\cg{add more on uncertainty quantification in conclusion}

%\rtwo{Other relevant topics related to inference in the high-dimensional regime, Bayesian approaches to neural networks, Issues related to generative models, etc., should be briefly mentioned as topics of interest and current research in the Conclusions section.}

We conclude our discussion by emphasizing that uncertainty has multiple sources and ignoring it can have severe consequences on the validity of trained supervised machine learning models. We looked at this question from a mainly statistical point of view and aimed to relate the discussion to traditional fields in statistics. Thereby, we also hope to contribute to the development of the basic science underpinnings of data-centric machine learning.

The key takeaways are: First, basic classical probability theory is in our view a centrepiece to describe and define aleatoric uncertainty. Second, the common definition of epistemic uncertainty as the ``remaining uncertainty'' is vague and therefore of limited value. The bias-variance decomposition provides a possible mathematical formalization of the idea. We note, however, that the additive formula ``aleatoric uncertainty + epistemic uncertainty = total uncertainty'' is not universally valid. 
Third, contrary to claims from machine learning about the flexibility of deep learning models, we argue that model uncertainty remains an issue if the available data are not suitable to infer the desired relationship between variables. Omitted variables, measurement errors, and non- \textit{i.i.d.} data are our particular concern. Fourth, we point to additional sources of uncertainty: the ones summarized in the Total Survey Error (TSE) framework, the challenges around missing data, or when deploying machine learning solutions in changing environments. Readers will easily identify others. All this shows that a simple decomposition of uncertainty into aleatoric and epistemic does not do justice to a much more complex constellation with multiple sources of uncertainty. It is crucial to explore how various sources of uncertainty relate to aleatoric and epistemic uncertainty. For instance, data uncertainty and aleatoric uncertainty cannot simply be equated. 
Similarly, we believe that it is ultimately not expedient to simply define aleatoric uncertainty as ``that which is irreducible'', at least without specifying exactly what (ir)reducible means. While increasing sample size ceteris paribus will not reduce aleatoric uncertainty, other characteristics of data do: we highlighted, e.g., the collection of additional features, features of higher quality (fewer errors or less item missingness), selectivity, and representativity. Conversely, the term irreducibility, with its implied focus on sample size, may lead us to forget that such aspects of data can actually be designed. 

We did not put emphasis on quantifying uncertainty. This apparently would be the next step. 
Methods to quantify uncertainty include Gaussian processes \citep{rasmussen2006GaussianProcessesMachine}, ensemble methods \citep{lakshminarayanan2017SimpleScalablePredictive}, and posterior inference on Bayesian neural networks via Hamiltonian Monte Carlo or stochastic variational inference \citep{psaros2023UncertaintyQuantificationScientific}. 
However, before quantifying uncertainty, it is necessary to allocate the possible sources.

Moreover, we remained with the probability paradigm \eqref{eq:datagen}, which itself is a \textit{model}. We refer to \citet{walley2000towards} or \citet{augustin2014introduction} for an alternative utilizing imprecise probabilities. 

Furthermore, we also did not focus on (algorithmic) fairness issues in machine learning which are connected to uncertainty as well as bias and its sources. Though this topic is of central importance and most topics in Sections \ref{sec:role.of.data} %\ref{sec:estimation_model_uncertainty} 
and \ref{sec:data-uncertainty} can be extended to discuss fairness, we considered it to be beyond the scope of this paper. % to relate fairness to uncertainty. 
We refer to \citet{mehrabi2021survey} for a survey of this topic, see also \citet{bothmann2022fairness}.

%%%%%%%%%%%%%%%%%%%%%%%%%%%%%%%%%%%%%%%%%%%%%%
%% Single Appendix:                         %%
%%%%%%%%%%%%%%%%%%%%%%%%%%%%%%%%%%%%%%%%%%%%%%
%\begin{appendix}
%\section*{???}%% if no title is needed, leave empty \section*{}.
%\end{appendix}
%%%%%%%%%%%%%%%%%%%%%%%%%%%%%%%%%%%%%%%%%%%%%%
%% Multiple Appendixes:                     %%
%%%%%%%%%%%%%%%%%%%%%%%%%%%%%%%%%%%%%%%%%%%%%%
%\begin{appendix}
%\section{???}
%
%\section{???}
%
%\end{appendix}
\begin{appendix}
%\appendix
%\section{Appendix (probably: Supplementary Material)}

% see https://tex.stackexchange.com/questions/5314/equations-in-section-heading-title\textbf{} for why there was a warning
\section{Consistency of Maximum Likelihood Estimate} 
In the following, we sketch the proof that the maximum likelihood estimate is consistent even for misspecified models \citep[see also][p. 234]{kauermann2021statistical}.

\label{appendix:convergence_MLE}
Let $\theta_0$ be implicitly defined through (\ref{eq:theta0}) and let the maximum likelihood be defined through 
\[
0=  \frac{\partial l (\hat{\theta})}{\partial \theta}.
\]
Then, simple Taylor Series expansion yields
\begin{align*}
 0 &= \frac{\partial l ({\theta_0})}{\partial \theta} + \frac{\partial^2 l (\tilde{\theta})}{(\partial \theta)^2} (\hat{\theta} - \theta_0) \\
\iff  \hat{\theta} - \theta_0 &= \left[ - \frac{\partial^2 l (\tilde{\theta})}{(\partial \theta)^2}\right]^{-1} \frac{\partial l ({\theta_0})}{\partial \theta}.
\end{align*}
where $\tilde{\theta}$ lies between $\theta_0$ and $\hat{\theta}$, using the mean value theorem of differentiation.

With the central limit theorem and with the definition of $\theta_0$ we get for misspecified models
\[
\frac{\partial l ({\theta_0})}{\partial \theta} \overset{a}{\sim} N\left(0, V(\theta_0)\right)
\]
with the variance $V(\theta_0) = Var(\frac{\partial l (\theta_0)}{\partial \theta})$, which could be estimated. Since 
$\tilde{\theta} \rightarrow \theta_0$
and applying standard asymptotic arguments as in \citet{mccullagh1987TensorMethodsStatisticsa} we obtain 
\[
 \hat{\theta} - \theta_0 \overset{a}{\sim} N\left(0, I^{-1}(\theta_0)V(\theta_0)I^{-1}(\theta_0)\right)
\]
where $I(\theta_0)$ is the Fisher information. This in turn proves consistency of $\hat{\theta}$ with respect to $\theta_0$.

\section{Omitted Variables% and the Simpson Paradox revisited
}
\label{Appendix:Omitted}\label{sec:app.omitted variables}

\newcommand{\ubar}[1]{\underset{\bar{}}{#1}}

%\subsection{Proof Sketches}\label{subsec:app.omitted.variables.math}
%\subsubsection*{Proof Sketch for \eqref{eq:varincrease}}\label{subsubsec:app.omitted.variables.math.proof.of.eq}
%to do.

%\begin{align}
%\nonumber
%    Var(Y|x) & =    
%  \int_{z \in {\cal Z}}  Var(Y| x, z)   f_{Z|X}(z|x) \: dz \\
%  & +  \int_{z \in {\cal Z}}
%  \left(E(Y|x,z) - E(Y|x)\right)^2 f_{Z|X}(z|x) dz        \notag \\
%  & = E_{Z|x}\left( Var(Y| x, z)\right) +  E_{Z|x}\left( bias(x,z)^2 \right).
%  \tag{\ref*{eq:varincrease}}
%\end{align}

\subsection*{Comparison of $Var_{Y|X,Z}(Y|x,z)$ and $Var_{Y|X}(Y|x)$}
\label{subsubsec:app.omitted variables.math.compare.variances}
Let $x$ be an arbitrary value with $f_{Z|X}(z|x) > 0$ for some $z$. 
%  $Var(Y|x)  =    
%  \int_{z \in {\cal Z}}  Var(Y| x, z)   
%  \left(E(Y|x,z) - E(Y|x)\right)^2 f_{Z|X}(z|x) dz  
%   = E_{Z|x}\left( Var(Y| x, z)\right) +  E_{Z|x}\left( bias(x,z)^2 \right)$

\paragraph*{\textit{Case 1:}}
Let the conditional variance in the full model be constant for all $z$ with $f_{Z|X}(z|x) > 0$ and let it exist. 
Then, 
\[
E_{Z|X}\left( Var(Y| x, z)\right) = Var_{Y|X,Z}(Y| x, z)
\]
for all $z$ with $f_{Z|X}(z|x) > 0$. As the expected squared bias must be non-negative, it follows from \eqref{eq:varincrease} that $Var_{Y|X,Z}(Y| x, z) \le Var_{Y|X}(Y|x)$ for all $x$ and all $z$ with $f_{Z|X}(z|x) > 0$. Further, 
\[
Var_{Y|X,Z}(Y| x, z) < Var_{Y|X}(Y|x)
\]
for all $z$ with $f_{Z|X}(z|x) > 0$ among those $x$ with positive expected squared bias.
%$Var(Y| x, z) \eqqcolon v(x)$ for all $z$. 

\paragraph*{\textit{Case 2:}}
Contrary to \textit{case 1}, let the conditional variance in the full model now not be constant for all $z$ with $f_{Z|X}(z|x) > 0$ and let it exist. 

Then, because of the monotonicity of the expected value operator, there exist $\bar{z}$ and $\ubar{z}$ with $f_{Z|X}(\bar{z}|x) > 0$ and $f_{Z|X}(\ubar{z}|x) > 0$ such that 
\[
Var_{Y|X,Z}(Y| x, \bar{z})  >  E_{Z|X}\left( Var_{Y|X,Z}(Y| x, z)\right)
\]
and also
\[
Var_{Y|X,Z}(Y| x, \ubar{z}) <  E_{Z|X}\left( Var_{Y|X,Z}(Y| x, z)\right).
\]
By plugging the second inequality into \eqref{eq:varincrease} and minding the non-negative expected squared bias, we see that 
\begin{align*}
    Var_{Y|X}(Y|x) & >   Var_{Y|X,Z}(Y| x, \ubar{z}) + E_{Z|x}\left( bias(x,z)^2 \right) \\
&\ge Var_{Y|X,Z}(Y| x, \ubar{z}).
\end{align*}

That is, for all $x$ there exist $\ubar{z}$, with $f_{Z|X}(\ubar{z}|x) > 0$, for which the conditional variance in the full model is smaller than the conditional variance in the omitted variable model.

For $\bar{z}$, we get 
\[
Var_{Y|X}(Y|x) <  Var_{Y|X,Z}(Y| x, \bar{z}) + E_{Z|X}\left( bias(x,z)^2 \right).
\]
Thus, in order for $Var_{Y|X}(Y|x) <  Var_{Y|X,Z}(Y| x, \bar{z})$ to be fulfilled, the expected squared bias must be small enough relative to the difference in the conditional variances: e.g., the bias may be zero. That such values $\bar{z}$ can exist but do not have to, is best demonstrated by two examples. Consider a linear true relationship, 
\[
Y = \beta_0 + \beta_X X + \beta_Z Z + \epsilon, \;\; \epsilon \sim N\left(0, Var(Y| x, z) \right).
\]
Let $Z \in \lbrace 0, 1 \rbrace$ be binary and $\Prob(Z=1|x),\,\Prob(Z=0|x) >0$.

First, if $\beta_Z = 0$, then $bias(x,z)=0 \, \forall x,z$. Letting $Var(Y| x, z=0) < Var(Y| x, z=1)$, it follows that
\begin{align*}
    Var(Y| x, z=1) >& \Prob(Z=0|x)Var(Y| x, z = 0) \\
    &+ \Prob(Z=1|x)Var(Y| x, z = 1) \\
    &= E_{Z|x}\left( Var(Y| x, z)\right) + 0 = Var(Y|x)
\end{align*}
and the desired $\bar{z}$ exists: $\bar{z}=1$.% This example corresponds to Scenario XXX below.

\phantomsection\label{app.omitted.linear.model.example}
Second, let $X$ and $Z$ be independent. Further, to make calculations simpler, let $\beta_Z =1$, $\Prob(Z=1) = \Prob(Z=0) = 0.5$, and $Var(x,z=0) =0.1 < 0.2 = Var(x,z=1)$. Thus, $E_{Z|x}\left( Var(Y| x, z)\right) = 0.15$. Further, for the omitted variable model, $Y = \tilde{\beta}_0 + \tilde{\beta}_X X + \tilde{\epsilon}$, it is established (\citealp[Ch.~3.4.1]{fahrmeir.et.al.2022.regression}; \citealp[Ch.~4.3.1]{wooldridge.2005.econometrics}) that then $\tilde{\beta}_0 = \beta_0 + \Prob(Z=1) \beta_Z$ and $\tilde{\beta}_X = \beta_X$. Thus, the absolute bias is constant, $|bias(x,z)|=0.5 |\beta_Z|$, and hence $E_{Z|x}\left( bias(x,z)^2 \right) = 0.25 \beta_Z^2 = 0.25$. Therefore, $0.1 \le Var(Y|x,z) \le 0.2 < 0.4 = 0.15 + 0.25 = Var(Y|x)$: the conditional variance for all $z$ is smaller in the full model than in the omitted variable model, i.e., the desired $\bar{z}$ does not exist.
% $ $ \int_{z \in {\cal Z}}  Var(Y| x, z)   
%  \left(E(Y|x,z) - E(Y|x)\right)^2 f_{Z|X}(z|x) dz  
%   = E_{Z|x}\left( Var(Y| x, z)\right) +  E_{Z|x}\left( bias(x,z)^2 \right)$
%\begin{align}
%\nonumber
%    Var(Y|x) & =    
%  \int_{z \in {\cal Z}}  Var(Y| x, z)   f_{Z|X}(z|x) \: dz \\
%  & +  \int_{z \in {\cal Z}}
%  \left(E(Y|x,z) - E(Y|x)\right)^2 f_{Z|X}(z|x) dz        \notag \\
%  & = E_{Z|x}\left( Var(Y| x, z)\right) +  E_{Z|x}\left( bias(x,z)^2 \right).
%  \tag{\ref*{eq:varincrease}}
%\end{align}

%\section{Measurement Error in  $X$} section sollte gleich heissen wie die section im Haupttext.
\section{Errors in $X$}
\subsection*{Discussion of the Assumption of Independence of $Y$ and $X$ given $Z$}\label{sec:app.errors.in.x.discussion.independence} 
If $Y$ is not independent of $X$ given $Z$, then $X$ contains information about $Y$ that is not contained in the error-free features $Z$. We see two possible reasons for this scenario. 

First, $X$ (also) contains omitted variables, i.e., features that influence $Y$ and that are not part of $Z$. However, that is not how we constructed $Z$ in our setup. In addition, we do not want to add an omitted variable problem on top of a measurement error problem in this section.  

Second, $Y$ may influence $X$, e.g., when $Y$ is temporally situated before $X$. In such a case, there would be an additional arrow $X \longleftarrow Y$ in Figure~\ref{fig:errorx}. 

We note that traditional errors-in-variables models make this independence assumption. It may not be satisfied in every setting.

%%%%%%%%%%%%%%%%%%%%%%%%%%%%%%%%%%%%%%%%%%%%%%%%%%%%%%%%%%%%%%%%%%%%%%%%%%%%%%
%\section{Appendix for Section~\ref*{subsec:errors.in.y},~\nameref*{subsec:errors.in.y}}%\newpage
%\pagebreak
%\section{Measurement Error in $Y$} section sollte gleich heissen wie die section im Haupttext.
\section{Errors in  $Y$}
\label{sec:app.errors.in.y}

\subsection{Unbiased Classification}	\label{subsec:app.errors.in.y.bias}
We consider multi-class prediction with true values $Z$ and observed, error-prone values $Y$.

The sum in \eqref{eq:true_y_given_x_discrete} can be split: either $Z = \upsilon$ or $Z \neq \upsilon$. 
% ${\cal Z} = \left\lbrace \upsilon \right\rbrace \,\dot{\cup}\,\, $ and use the complementarity of $y$ and $y^c$ also on the probability 
Next, we use that, since the probabilities always add to 1,
$\Prob_{Y|X,Z} (Y= \upsilon| X=x,Z=z) = 1 - \Prob_{Y|X,Z} (Y\neq \upsilon| X=x,Z=z)$ for all values of $z$, and hence also for $z = \upsilon$.
%$\Prob_{Y|X,Z} (Y= \upsilon| X=x,Z= \upsilon) = 1 - \Prob_{Y|X,Z} (Y\neq \upsilon| X=x,Z=\upsilon)$:
\begin{align}
&	\Prob_{Y|X}(Y= \upsilon|X=x) \\
&	= \sum_{z \in {\cal Z}}  \Prob_{Y|X,Z} (Y= \upsilon|X=x,Z=z) \Prob_{Z|X}(Z=z|X=x) \nonumber\\
&	=  \Prob_{Y|X,Z} (Y= \upsilon|X=x,Z= \upsilon) \Prob_{Z|X}(Z= \upsilon|X=x) \nonumber\\
&	\quad  + \Prob_{Y|X,Z} (Y= \upsilon| X=x,Z\neq \upsilon) \Prob_{Z|X}(Z\neq \upsilon|X=x) \nonumber\\
	%\overset{(a)}{=}
&	= \left( 1 - \Prob_{Y|X,Z} (Y\neq \upsilon| X=x,Z= \upsilon) \right)  \Prob_{Z|X}(Z= \upsilon|X=x) \nonumber\\
&   \quad + \Prob_{Y|X,Z} (Y= \upsilon| X=x,Z\neq \upsilon) \Prob_{Z|X}(Z\neq \upsilon|X=x) \nonumber\\ 
&	= \Prob_{Z|X}(Z= \upsilon|X=x) \nonumber\\
&	 \quad - \underbrace{\Prob_{Y|X,Z} (Y\neq \upsilon| X=x,Z= \upsilon) \Prob_{Z|X}(Z= \upsilon|X=x)}_{\Prob_{Y,Z|X} (Y\neq \upsilon, Z = \upsilon| X=x)} \nonumber\\
&	\quad  + \underbrace{\Prob_{Y|X,Z} (Y= \upsilon| X=x,Z\neq \upsilon) \Prob_{Z|X}(Z\neq \upsilon|X=x)}_{\Prob_{Y,Z|X} (Y= \upsilon, Z \neq \upsilon| X=x)}\nonumber\\
&	= \Prob_{Z|X}(Z= \upsilon|X=x) \nonumber\\
&	\quad  + \Prob_{Y,Z|X} (Y= \upsilon, Z \neq \upsilon| X=x)  \nonumber\\
&   \quad  - \Prob_{Y,Z|X} (Y\neq \upsilon, Z = \upsilon| X=x) \nonumber
	 \label{eq:app.true.false.Z.Y.bias.multiclass}
\end{align}

For %the special case 
binary classification, simply replace ``$=\upsilon$'' with ``$=1$'' and replace ``$\neq \upsilon$'' with ``$=0$'' everywhere to arrive at equation \eqref{eq:true.false.Z.Y.bias}. 

%\eqref{eq:true.false.Z.Y.bias} generalizes to multi-class prediction (Appendix~\ref{subsec:app.errors.in.y.bias}): for every value $\upsilon$ in the support of $Y$ and of $Z$,% ${\cal Z} = {\cal Y}$, 
%\begin{align}
%	\Prob_{Y|X}(Y=\upsilon|X=x) = \Prob_{Z|X}(Z=\upsilon|X=x) + &&\nonumber %\\
%	\quad {} \Prob_{Y,Z|X} (\underbrace{Y=\upsilon, Z \neq \upsilon}_{\text{%cond. probability of 
%			false $Y=\upsilon$}}| X=x) - \Prob_{Y,Z|X} (\underbrace{Y\neq \upsilon, Z=\upsilon}_{\text{%cond. probability of 
%			false $Y\neq \upsilon$}}| X=x) \, .
%	\label{eq:true.false.Z.Y.bias.multiclass}
%\end{align}	
%\in {\cal Y} \setminus \left \lbrace \upsilon \right\rbrace, Z=\upsilon
For unbiased binary classification, \eqref{eq:true.false.Z.Y.bias} needs to hold for every $x$. For unbiased multi-class classification, \eqref{eq:app.true.false.Z.Y.bias.multiclass} must hold for every $x$ and also for every $\upsilon$ in the support: the number of necessary assumptions is thus multiplied with a factor equal to the size of the support minus 1; as the probabilities sum to 1, the last probability is already determined. This is also why it is sufficient to write \eqref{eq:true.false.Z.Y.bias} only for one value, $\upsilon = 1$.

\subsection{The Role of Conditional Error Equality}\label{subsec:app.errors.in.y.cond.error}	
Assume that the conditional error probabilities are identical, 
\begin{align}
  \tag{\ref{eq:true.false.Z.Y.bias.same.cond.error}} %same eq number as in the main text (errors-in-y)
	&\Prob_{Y|X,Z} (\underbrace{Y=1|X=x, Z=0}_{\text{false $Y=1$}}) =& \\
	&\Prob_{Y|X,Z} (\underbrace{Y=0|X=x, Z=1}_{\text{false $Y=0$}})  \eqqcolon c(x).& \nonumber \notag 
\end{align}
%\begin{align}
%	\Prob_{Y|X,Z} (\underbrace{Y= \upsilon|X=x, Z\neq \upsilon}_{\text{false $Y= %\upsilon$}}) =
%	\Prob_{Y|X,Z} (\underbrace{Y\neq \upsilon|X=x, Z= \upsilon}_{\text{false $Y\neq \upsilon$}}) =: c(x)
%	\label{eq:true.false.Z.Y.bias.multiclass.conditional.errors.equal}
%\end{align}
The bias does not vanish unless the marginal conditional probabilities  
$\Prob_{Z|X}(Z=0|X=x)$ and $\Prob_{Z|X}(Z=1|X=x)$
are also identical:
\phantomsection\label{sec:appendix.true.false.Z.Y.bias.conditiona}
\begin{align}
	\label{eq:true.false.Z.Y.bias.conditional}
&	bias_{Y|X}(Y=1|X=x)  \\
&	=  \Prob_{Y,Z|X} (\underbrace{Y=1, Z=0}_{\text{%cond. probability of 
			false $Y=1$}}| X=x) \nonumber\\
&   \quad   - \Prob_{Y,Z|X} (\underbrace{Y=0, Z=1}_{\text{%cond. probability of 
			false $Y=0$}}| X=x) \nonumber\\
&	=  \Prob_{Y|X,Z} (Y=1|X=x, Z=0)\Prob_{Z|X}(Z=0|X=x) \nonumber \\
&	\quad - \Prob_{Y|X,Z} (Y=0|X=x, Z=1)\Prob_{Z|X}(Z=1|X=x) \nonumber \\
&	= 	c(x) \left(	\Prob_{Z|X}(Z=0|X=x) - \Prob_{Z|X}(Z=1|X=x) \right).	\nonumber
\end{align}
%\begin{align}
%	bias&_{Y|X}(Y=\upsilon|X=x)  \nonumber\\
%	= & \Prob_{Y,Z|X} (\underbrace{Y=\upsilon, Z\neq \upsilon}_{\text{%cond. probability of 
		%			false $Y=\upsilon$}}| X=x) - \Prob_{Y,Z|X} (\underbrace{Y\neq \upsilon, Z=\upsilon}_{\text{%cond. probability of 
		%			false $Y\neq \upsilon$}}| X=x) \nonumber\\
%	= & \Prob_{Y|X,Z} (Y=\upsilon|X=x, Z\neq \upsilon)\Prob_{Z|X}(Z\neq \upsilon|X=x) \nonumber \\
%	& - \Prob_{Y|X,Z} (Y\neq \upsilon|X=x, Z=\upsilon)\Prob_{Z|X}(Z=\upsilon|X=x) \nonumber \\
%	= &	c(x) \left(	\Prob_{Z|X}(Z\neq \upsilon|X=x) - \Prob_{Z|X}(Z=\upsilon|X=x) \right)	\, .
%	\label{eq:true.false.Z.Y.bias.multiclass.conditional}
%\end{align}
With equal and positive conditional error probability $c(x)>0$, unbiasedness is only possible iff the term in parentheses is zero, i.e., iff $\Prob_{Z|X}(Z=0|X=x) = \Prob_{Z|X}(Z=1|X=x)$. % If we add to that $\Prob_{Z|X}(Z=1|X=x)$ on both sides and use that probabilities always sum to 1, $\Prob_{Z|X}(Z=0|X=x)+\Prob_{Z|X}(Z=1|X=x) = 1$, we see what is required:
As these two complementary probabilities must also sum to 1, we have
\begin{align}
	\Prob_{Z|X}(Z=1|X=x) = 0.5 = \Prob_{Z|X}(Z=0|X=x),
	\tag{\ref{eq:true.false.Z.Y.bias.conditional.all.0.5}}
\end{align}
%\begin{align}
%	\Prob_{Z|X}(Z=1|X=x) = 0.5 & \text{ and } \Prob_{Z|X}(Z=0|X=x) = 0.5 \, ,
%	\label{eq:eq:true.false.Z.Y.bias.multiclass.conditional.all.0.5}
%\end{align}
%\begin{align}
%	\Prob_{Z|X}(Z=\upsilon|X=x) = 0.5 & \text{ and } \Prob_{Z|X}(Z\neq \upsilon|X=x) = 0.5 \, .
%	\label{eq:eq:true.false.Z.Y.bias.multiclass.conditional.all.0.5}
%\end{align}
% $\Prob_{Z|X}(Z\neq \upsilon|X=x) = \Prob_{Z|X}(Z=\upsilon|X=x) = 0.5$ is required.
i.e., the binary classification problem must be exactly balanced for the bias to vanish. As the two conditional marginal probabilities in \eqref{eq:true.false.Z.Y.bias.conditional.all.0.5} must be equal to 0.5 for all $x$, this implies independence of $X$. 
%In conclusion, positive and identical conditional error probabilities lead to bias of $\Prob_{Y|X}$ for $\Prob_{Z|X}$ in binary classification unless the classes are always perfectly balanced and $X$ is irrelevant -- but why do classification in that scenario? Equality of conditional error probabilities is simply not very helpful.

%%%%%%%%%%%%%%%%%%%%%%%%%%%%%%%%%%%%%%%%%%%%%%%%%%%%%%%%%%%%
%%%%%%%%%%%%%%%%%%%%%%%%%%%%%%%%%%%%%%%%%%%%%%%%%%%%%%%%%%%%
\section{Missing Data}
\label{sec:app.missing.data}

\subsection{Some Formal Background}
\phantomsection\label{sec:app.proof.of.eq.missing.data.mechanism}

\subsection*{Proof Sketch for \eqref{eq:missing.data.mechanism}}
We use the definition of a conditional density/probability repeatedly:
\begin{align*}
f_{Y|X,R}(y|x,R=1) 
    = & \frac{f_{X,Y,R}(x,y,R=1)}{f_{X,R}(x,R=1)} \\
    = & \frac{f_{R|X,Y}(R=1 | x,y) f_{X,Y}(x,y)}{f_{R|X}(R=1|x)f_X(x)} \\
    = & \frac{f_{R|X,Y}(R=1 | x,y)}{f_{R|X}(R=1|x)}  f_{Y|X}(y|x)  \\
    = & \frac{\Prob (R=1 | y , x )}{\Prob(R=1 | x )}  f_{Y|X}(y|x).
\end{align*}
The requirements $f_{R|X}(R=1|x) = \Prob(R=1 | x ) > 0$ and $f_X(x) > 0$ are naturally met since otherwise we could have never data for $x$. 

\subsection*{Equivalence of $R \not\!\perp\!\!\!\perp Y | X$ and $f_{Y|X,R}(y|x,R=1) = f_{Y|X,R}(y|x,R=0)$}
Analogously to \eqref{eq:missing.data.mechanism}, we get
\begin{align}
    \tag{\ref*{eq:missing.data.mechanism}'}
    \label{eq:missing.data.mechanism.prime}
    &f_{Y|X,R}(y|x,R=0) = \frac{\Prob (R=0 | y , x )}{\Prob(R=0 | x )} f_{Y|X}(y|x) \\ 
    &= \frac{1 - \Prob (R=1 | y , x )}{1 - \Prob(R=1 | x )} f_{Y|X}(y|x).\notag
 \end{align}
Assuming $f_{Y|X}(y|x)>0$, the equivalence of the conditional independence implied by \eqref{eq:missing.data.mechanism} and $f_{Y|X,R}(y|x,R=1) = f_{Y|X,R}(y|x,R=0)$ follows from \eqref{eq:missing.data.mechanism} and \eqref{eq:missing.data.mechanism.prime}:
% \begin{align*}
%     & f_{Y|X,R}(y|x,R=1)  &=& f_{Y|X,R}(y|x,R=0)& \notag\\
% \Leftrightarrow & 
%     \frac{\Prob (R=1 | y , x )}{\Prob(R=1 | x )}  &=&
%     \frac{1 - \Prob (R=1 | y , x )}{1 - \Prob(R=1 | x )}&  \notag\\
% %\Leftrightarrow & 
% %    \Prob (R=1 | y , x ) \left( 1 - \Prob(R=1 | x ) \right) &=&  \Prob(R=1 | x ) \left( 1 - \Prob (R=1 | y , x ) \right) &\notag\\
% \Leftrightarrow & 
%     \Prob (R=1 | y , x ) &=&  \Prob(R=1 | x )  &\notag\\
%  \end{align*}
\begin{align*}
    f_{Y|X,R}(y|x,R=1)  =& f_{Y|X,R}(y|x,R=0)& \Leftrightarrow \notag\\
    \frac{\Prob (R=1 | y , x )}{\Prob(R=1 | x )}  =&
    \frac{1 - \Prob (R=1 | y , x )}{1 - \Prob(R=1 | x )}&  \Leftrightarrow \notag\\
%\Leftrightarrow & 
%    \Prob (R=1 | y , x ) \left( 1 - \Prob(R=1 | x ) \right) &=&  \Prob(R=1 | x ) \left( 1 - \Prob (R=1 | y , x ) \right) &\notag\\
    \Prob (R=1 | y , x ) =&  \Prob(R=1 | x )  & 
 \end{align*}
 
\subsection*{Differences between Respondents, Nonrespondents, and the Population Average}
Using \eqref{eq:missing.data.mechanism} and \eqref{eq:missing.data.mechanism.prime}, it follows that the conditional independence of missingness and $Y$ given $X$ is sufficient for no difference between the conditional expectations of respondents and nonrespondents, respectively: 
\begin{align*}
     & E_{Y|X,R}(Y|x, R = 1) - E_{Y|X,R}(Y|x, R = 0) \notag\\
    =& \int y \left( f_{Y|X,R}(Y|x, R = 1) - f_{Y|X,R}(Y|x, R = 0) \right) dy \notag\\
    =& \int y f_{Y|X}\left( \frac{\Prob (R=1 | y , x )}{\Prob(R=1 | x )} - \frac{1 - \Prob (R=1 | y , x )}{1 - \Prob(R=1 | x )} \right) dy \notag\\
    =& \int y f_{Y|X}  \frac{\Prob (R=1 | y , x ) - \Prob(R=1 | x )}{\Prob(R=1 | x )\left(1 - \Prob(R=1 | x )\right)}  dy.%\notag\\
\end{align*}
Conditional independence is not necessary, however: in particular, $Y|x, R = 1$ and $Y|x, R = 0$ may have the same expectation, but one may have a larger variance. By \eqref{eq:bias.between.respondents.and.population.average}, this translates to the same statements about a difference between respondents and the population average or a difference between nonrespondents and the population average.

\subsection{Discussion of Missing At Random (MAR) Data}
\label{subsec:app.discussion.of.mar}
On the one hand, regardless of whether $Y$ is observed or not: as stated in the discussion of \eqref{eq:missing.data.mechanism}, if missingness depends on $Y$ given $X$, the bias factor is not equal to 1. On the other hand, if missingness is independent of $Y$ given $X$, $\Prob (R=1 | y , x ) = \Prob(R=1 | x )$, then by \eqref{eq:missing.data.mechanism} unbiasedness is already guaranteed; splitting $X= (X_{\text{miss}},X_{\text{obs}})$ %the features 
into a missing and an observed part %, $X= (X_{\text{miss}},X_{\text{obs}})$,
and imposing the stronger MAR assumption, $P(R|x_{\text{obs}},x_{\text{miss}}) = P(R|x_{\text{obs}})$,  provides no additional benefit. In sum, MAR is neither necessary nor sufficient. 

 %For instance, if in (\ref{eq:missing.data.mechanism}) missingness in $Y$ depends only on (fully observed) $X$, the bias factor equals 1. 
%If, however, missingness occurs also in parts of the input variables $X$, even in  the MAR setup, this implies a bias. -> Nein, das ist nicht richtig. Teile von X können fehlen und es ist kein Problem in MAR, solange die missingness nicht von diesen  fehlenden Teilen abhaengt, sondern nur von X_obs.
%Splitting $X= (X_{\text{miss}},X_{\text{obs}})$ %the features 
%into a missing and an observed part, %$X= (X_{\text{miss}},X_{\text{obs}})$,
%MAR for item nonresponse in $X$ means $P(R|y,x_{\text{obs}},x_{\text{miss}}) = P(R|y,x_{\text{obs}})$ which does not simplify the bias factor in (\ref{eq:missing.data.mechanism}). Mein: kein bias, also auch keine Imputation oder weighting nötig! In this case, one might use inverse probability weighting or multiple imputations to recover the desired  $f_{Y|X}$ despite missingness. 

\phantomsection\label{missing.data.auxiliary.data.w}
Furthermore, in predictive modelling, using $Y$ in an imputation model for missing feature values $X$ often constitutes impermissible target leakage \citep[Ch.~7.8.1]{ghani.schierholz.2020.machine.learning.ml.in.big.data.and.social.science.book}. (In traditional statistical analysis, i.e., in estimation instead of prediction, this is not a problem.) 
One way out are additional variables $W$, correlated with both, $R$ and $Y$ conditional on $X$, that break the conditional dependence of response and outcome: $\Prob (R=1 | y , x , w) = \Prob (R=1 | x_{\text{obs}}, w)$. Auxiliary information such as paradata \citep{kreuter.2013.paradata.book,schenk.reuss.2023.paradata.in.surveys} are an example for such $W$ that helps to recover $f_{Y|X}$, but would not be used as features $X$ in the prediction model. In fact, for \textit{unit} nonresponse the MAR concept makes only sense in the presence of such $W$.

\end{appendix}

\section*{Funding}
CG is supported by the DAAD programme Konrad Zuse Schools of Excellence in Artificial Intelligence, sponsored by the Federal Ministry of Education and Research. 
POS is supported in part by the Federal Statistical Office of Germany. MS has been partially funded by the Deutsche Forschungsgemeinschaft (DFG, German Research Foundation) as part of BERD@NFDI - grant number 460037581.

\bibliographystyle{abbrvnat} 
%\bibliography{references}  

%%%%%%%%%% Merge with supplemental materials %%%%%%%%%%

\clearpage
\begin{localsize}{11}

\newgeometry{width = 160mm, top = 35mm, bottom = 30mm, bindingoffset = 0mm}
\onecolumn

\setcounter{equation}{0}
\setcounter{figure}{0}
\setcounter{table}{0}
\setcounter{page}{1}

\renewcommand*{\thesection}{\arabic{section}}
\setcounter{section}{0}

\begin{center}

{\LARGE Supplement to ``Sources of Uncertainty in Supervised Machine Learning - A Statisticians' View''} \\[1cm]
\large
Cornelia Gruber\footnote{cornelia.gruber@stat.uni-muenchen.de}, Patrick Oliver Schenk, Malte Schierholz, Frauke Kreuter, and G{\"{o}}ran Kauermann \\[.4cm]
Department of Statistics, Ludwig-Maximilians-University Munich, Ludwigstr. 33,  80539 Munich, Germany \\[1cm]
\end{center}

\section{Omitted Variables}
\label{sec:supp.omitted}

\subsection{Relation between Variance Heterogeneity and Bias for Binary $Y$}\label{subsubsec:app.omitted variables.math.binary.Y}
\label{subsec:var.heterogeneity.and.bias.in.binary.y}
%Kann evtl. auch ins Supplementary Material.\\
%\gk{JA, oder ganz raus. Das ist schon sehr speziell. Ich schlage vor raus}
%\ps{Nachdem im Text darauf Bezug genommen wird und daraus eine Schlussfolgerung abgeleitet wird, wuerde ich sagen, SM ist ok. SM ist ja irgendwo "gratis".}
The statement to verify is that generally, the omitted variable model suffers either from both, neglected variance heterogeneity and bias for some $z$, or from neither. 

Let $z_1, z_2$ be two distinct values. Denote $p_1 \coloneqq \Prob(y|x,z_1)$ and $p_2 \coloneqq \Prob(y|x,z_2)$. For binary variables $Y$, it is well known that $E(Y|x,z)=\Prob(Y|x,z)$ and $Var(Y|x,z)=\Prob(Y|x,z) \left(1- \Prob(Y|x,z) \right)$.

Before starting, we consider when $Var(Y|x,z_1) = Var(Y|x,z_2)$:
 \begin{align*}
      Var(Y|x,z_1) = &   Var(Y|x,z_2)     & \Leftrightarrow \\
      p_1 (1-p_1)  = & p_2 (1-p_2)       & \Leftrightarrow \\
      p_1 - p_1^2  = & p_2 - p_2^2       & \Leftrightarrow \\
      p_1 - p_2    = & p_1^2 - p_2^2     & \Leftrightarrow \\
      p_1 - p_2    = & (p_1 + p_2) (p_1 - p_2)   %& \Leftrightarrow \\
 \end{align*}
As a quadratic equation, this has two solutions. First, if $p_1=p_2$, then both sides are equal to zero. Second, if $p_1 \neq p_2$, we have $p_1 = 1-p_2$. 

\textit{Variance Heterogeneity $\Rightarrow$ Bias:\\}
If variances are not homogeneous, there exist $z_1, z_2$, with $z_1 \neq z_2$ and $f_{Y|X,Z}(y|x,z_1), f_{Y|X,Z}(y|x,z_2) > 0$, such that $Var(Y|x,z_1) \neq  Var(Y|x,z_2)$. From above, we know that then $p_2 \notin \lbrace p_1, 1-p_1\rbrace$, and thus $E(Y|x,z_1)=p_1 \neq p_2 = E(Y|x,z_2)$. Therefore, $E(Y|x,z_1) \neq E(Y|x)$, $E(Y|x,z_2) \neq E(Y|x)$, or both, i.e., bias.

\textit{Bias $\Rightarrow$ Variance Heterogeneity:\\}
By bias we mean that there exist $z_1,z_2$, with $z_1 \neq z_2$ and $f_{Y|X,Z}(y|x,z_1), f_{Y|X,Z}(y|x,z_2) > 0$, such that $E(Y|x,z_1) < E(Y|x)$ and $E(Y|x,z_2) > E(Y|x)$. Thus, $E(Y|x,z_1)=p_1 \neq p_2 = E(Y|x,z_2)$. From above, we know that then $Var(Y|x,z_1) \neq   Var(Y|x,z_2) $, unless $p_1 \neq 1 - p_2$. This sole exception is to what the qualifier ``generally'' refers in our statement.

%Kurz für Malte die Idee:\\
%\begin{itemize}
%    \item wenn zwei Verteilungen zwei verschiedene Erwartungswerte haben $E_1 = p_1 \neq p_2 = E_2$ , dann deshalb weil die conditional probabilities $p_1 = \Prob(y|x,z_1)$ und $p_2 = \Prob(y|x,z_2)$ verschieden sind. Dann sind i.A. auch die Varianzen verschieden, $Var_1 = p_1 (1-p_1) \neq p_2 (1-p_2) = Var_2$; "i.A." wegen der Ausnahme $p_1 = 1-p_2$. Das laesst sich einfach zeigen, ist ne quadratische Gleichung.
%    \item Und umgekehrt wenn $Var_1 \neq Var_2$, dann sind auch $E_1 = p_1 \ neq p_2 = E_2$ die Erwartungswerte verschieden.
%\end{itemize}
%Liegt daran, dass es keinen separaten Streuungs"parameter" gibt, sondern nur einen "Parameter" für die Verteilung eines binären $Y$.

%\subsection{Scenarios in a Linear Model}\label{subsec:app.omitted.variables.scenarios}
%\pat{Ich sehe den Mehrwert nicht, den dieser Appendix hat. Das relevante ist im Haupttext. Und es wäre auch eine sehr ungleiche Ausgestaltung, hier mit linearen Modellen zu wuchern, während man in anderen Abschnitten sehr sparsam ist. Ausserdem wäre es ein grosser Zeitaufwand, den ich nicht gerechtfertigt sehe.}

\subsection{Marginal Effects and Simpson's Paradox}
\label{subsec:simpsons.paradox}
 Let us now consider the marginal effect of $X$ on $Y$. One may argue that marginal effects are much more important in causal inference than in prediction. However, marginal effects are an important tool in the increasing push to make (black box) machine learning models more interpretable and, often, the derived marginal effects are taken as glimpses of the true nature of the world instead of as mere descriptions of the learned model \citep{molnar.2020.interpretable.ml}. 
  %This may be less of a concern if the focus is exclusively on prediction, but if any interpretation is extracted from the trained model, one is exposed to the risk of misinterpretation due to omitted variables. 
More importantly for the topic of our paper, we show that even strong assumptions about the marginal effects of $X$ and about the relationship of $X$ and $Z$ are not enough for an omitted variable to be ignorable for uncertainty quantification in prediction. %We discuss the general results below and walk through them with a concrete example in Appendix~\ref{subsec:app.omitted.variables.scenarios}. 

Small changes in the input features $X$ lead to changes in the output according to 
\begin{align}
    \label{eq:expmissingz}
    \frac{\partial E(Y|x)}{\partial x} 
        =  \int_{z \in {\cal Z}} \frac{\partial E(Y | x, z)}{\partial x} f_{Z|X}(z|x) \: dz +  
          \int_{z \in {\cal Z}} E(Y | x, z) \frac{\partial f_{Z|X}(z|x)}{\partial x} \: dz.           
\end{align}
First, we notice that for the marginal effect of $X$ to be the same in the full model $Y|x,z$ and in the omitted variable model $Y|x$, it is not sufficient for the marginal effect %$\partial E(Y|x,z)/(\partial x)$
$\frac{\partial E(Y|x,z)}{\partial x}$ to be independent of $Z$ as there is a second term on the right-hand side of \eqref{eq:expmissingz}. Imagine a point $(x_0,z_0)$ for which the full and the omitted variable model make the same prediction. However, because the marginal effect in the omitted variable model is biased for the true marginal effect, it is not possible that the omitted variable model yields unbiased predictions for all $z$. Uncertainty measures such as prediction intervals that build on the expected conditional value of $Y$ are thus negatively affected. 
%Hence, even when the marginal effect is not modified by $Z$, the predictions are generally biased by omission of $Z$ and, thereby, uncertainty quantifications that build on the predicted mean are affected as well. \pat{to do an mich: eigenen Text verbessern.}

Second, we find that if $X$ and $Z$ are dependent then even the sign of the derivative can change. %For instance, when $\frac{\partial E(Y | x, z)}{\partial x} >0  \, \forall z$ and $E(Y | x, z)>0$, then a negative relationship between $X$ and $Z$, i.e., $\frac{\partial E(Y | x)}{\partial x} <0$ , then $\frac{\partial E(Y | x)}{\partial x} <0$
For instance, even if 
$\frac{\partial E(Y | x, z)}{\partial x} >0  \, \forall z$ we may find  $\frac{\partial E(Y | x)}{\partial x} <0$, particularly when $X$ and $Z$ are negatively related. This is typically known as Simpson's paradox.%, which is well known in statistics (see, e.g., \citealt{wagner1982simpson}) but less focused on in  machine learning (but see, e.g., \citealt{sharma2022detecting}). However, discussions of Simpson's paradox are concerned with the (parametric) marginal effects themselves and not with predictions and the quantification of their uncertainty. 

Third, for the marginal effect of $X$ to be the same in the full and in the omitted variable model, one requires strong assumptions: e.g., in addition to $Z$ not modifying the marginal effect of $X$, also that $X$ and $Z$ are independent so that $\frac{\partial f_{Z|X}(z|x)}{\partial x} = 0$. %and the second integral on the right-hand side of \eqref{eq:expmissingz} vanishes.
However, there may still be bias in the predictions so that prediction intervals will not be centred properly: 
%e.g., in the linear model example of Appendix~\ref{app.omitted.linear.model.example},
e.g., in a linear model example,
while the marginal effect is unaffected, the intercept is biased and so are, therefore, predictions. %(Scenario a/b in Appendix~\ref{subsec:app.omitted.variables.scenarios}) \highlight{TODO fix ref to appendix}.
According to (22) in the main paper, the bias also means that aleatoric uncertainty in the omitted variable model is higher than in the full model. Thus, even under these strong assumptions, $Z$ is not ignorable for uncertainty quantification.

\section{Errors in $Y$}
\label{sec:supp.errors,in.y}

\subsection{Example}\label{subsec:app.errors.in.y.example}
\begin{table}[!htb]
	\caption{Imbalanced classification problem: identical conditional error probabilities of $c(x) = 0.10$ produce bias as $\Prob_{Y|X}(Y=1|X=x)=0.74 < 0.80 = \Prob_{Z|X}(Z=1|X=x)$. All probabilities are conditional on $X$. Within-cell values are joint probabilities of $Y$ and $Z$, i.e., $\Prob_{Y,Z|X}(Y=y,Z=z|X=x)$.
	}
	\label{tab:true.false.Z.Y.cond.error}
 \centering
	\begin{tabular}{ cc|c|c|c }
		& \multicolumn{1}{c}{} & \multicolumn{2}{c}{$Y$ (Observed)} \\
		& \multicolumn{1}{c}{}
		&  \multicolumn{1}{c}{1}
		& \multicolumn{1}{c}{0}
		& $\sum$\\
		\cline{3-4}
		\multirow{4}{*}{$Z$ (Truth)}  & \multirow{2}{*}{1} & 0.72  & 0.08 & \multirow{2}{*}{0.80} \\
		  &  & {(true 1's)}  & {(false 0's)} & \\
		
		%	  &   &    	  & \red{missed 1 (missed 1)}) \\
		\cline{3-4}
		& \multirow{2}{*}{0} & 0.02 & 0.18 & \multirow{2}{*}{0.20}	  \\
	 	& & {(false 1's)} & {(true 0's)} & 	  \\
		\cline{3-4}
		&\multicolumn{1}{c}{$\sum$} &\multicolumn{1}{c}{0.74}  &\multicolumn{1}{c}{0.26}  & \multicolumn{1}{c}{1}
	\end{tabular}
\end{table}
%\input{double.tables}

%2
\autoref{tab:true.false.Z.Y.cond.error}  presents an example of how equal conditional error probabilities ($c(x)=0.10$) in \emph{imbalanced} binary classification ($\Prob_{Z|X}(Z=1|X=x) = 0.80$) produce bias: there are $0.10 \cdot 80 = 8$ percentage points too few 1's observed (in the $Z=1$ row), but the only $0.10 \cdot 20 = 2$ percentage points of false 1's (in the $Z=0$ condition) cannot make up for this, so that $\Prob_{Y|X}(Y=1|X=x) = 0.74$ exhibits a downward bias of 6 percentage points for the true probability $\Prob_{Z|X}(Z=1|X=x) = 0.80$.

If we keep the imbalance ($\Prob_{Z|X}(=1|X=x) = 0.80$) and the conditional error probability of false 0's at 0.10 from Table~\ref{tab:true.false.Z.Y.cond.error} but want unbiasedness, then the conditional error probability of false 1's must be 0.40 (see~\ref{subsec:app.errors.in.y.cond.error.minority} for more). This illustrates what the formalism of requiring the two \emph{joint} error probabilities to be equal 
%in \eqref{eq:true.false.Z.Y.bias} 
does not convey on its own: for unbiasedness, the conditional error probability in the minority class must be $\frac{\Prob_{Z|X}(Z=1|X=x)}{\Prob_{Z|X}(Z=0|X=x)}$ times %\replace{as high as}
{that} in the majority class, which can be a very strong requirement unless, of course, the conditional error probability in the majority class is close to zero.

%So far in \ref{subsec:errors.in.y}, we have only focused on the bias caused by the data's deficiency ($Y$ instead of $Z$). This bias cannot be reduced at all by increasing sample size. Note that the two additional components in \eqref{eq:true.false.Z.Y.bias}, the two error probabilities, are also subject to random variation: in finite samples, the empirical error rates for each value of $X$ will vary around the respective error probabilities. This additional source of uncertainty, however, does decrease with increasing sample size.

%This subsection focused on bias for a simple reason: many classifiers, to the extent that they provide uncertainty measures at all, express the uncertainty about the outcome $y\in \left\lbrace 0, 1 \right\rbrace$ only in terms of the predicted probability of $y=1$ and do not have additional parameters, unlike predictions intervals in regression (see \eqref{eq:prediction.interval.linear.regression}). If a data deficiency such as errors in $Y$ compromises these predictions, self-assessed uncertainty quantifications are affected as well.

\subsection{Conditional Error Probability in the Minority Class}	\label{subsec:app.errors.in.y.cond.error.minority}
Without loss of generality, assume that $0$ represents the minority class, i.e., $\Prob_{Z|X}(Z=0|X=x) < \Prob_{Z|X}(Z=1|X=x)$. The conditions for unbiasedness can be found by setting $bias_{Y|X}(Y=1|X=x)$ equal to zero in %\eqref{eq:true.false.Z.Y.bias.conditional}
(43) in the main paper and solving for the (joint) error probability of false 1's:
\begin{align}
	\label{eq:true.false.Z.Y.bias.condition.for.error.prob}
\Prob_{Y|X,Z}  (\overbrace{Y=1|X=x, Z=0}^{\text{false 1's}}) = 
  \Prob_{Y|X,Z} (\overbrace{Y=0|X=x, Z=1}^{\text{false 0's}}) 
 \cdot \frac{\Prob_{Z|X}(Z=1|X=x)}{\Prob_{Z|X}(Z=0|X=x)} , 
\end{align}
i.e., the conditional error probability in the minority class must be inversely proportional to the odds of the majority class and thus must always be greater than the  conditional error probability in the majority class.

If $Z=1$ is $\frac{0.80}{0.20}=4$ times as likely as $Z=0$, as in the example from Table~\ref{tab:true.false.Z.Y.cond.error}, the conditional error probability %of false 1's 
in the minority class 
thus would need to be four times as high as the conditional error probability in the majority class. If the latter is a not-uncommon 0.10, as in Table~\ref{tab:true.false.Z.Y.cond.error}, the former would need to be a whopping 0.40 for unbiasedness -- close to coin-flipping territory.

\subsection{Comments on Regression}\label{subsec:app.errors.in.y.comments.on.regression}
%Classification is a natural example for this subsection. We note that for 
In regression with unbiased errors of a quantitative $Y$, as modelled with the error term $\epsilon$ having expected value zero in linear regression, there is higher epistemic uncertainty in the training of the model but no bias in the parameter estimates \citep[Ch.~15]{carroll.et.al.2006.measurement.error}. However, when our focus is on prediction, even something benign such as unrecognized heteroscedasticity, which is still unbiased in expectation, produces miscalibrated pointwise predictive uncertainty.

%%%%%%%%%%%%%%%%%%%%%%%%%%%%%%%%%%%%%%%%%%%%%%%%%%%%%%%%%%%%%%%%%%%%%%%%%%%%%%
\section{Missing Data}
\label{sec:supp.missing}

\subsection{Relevance of Efficiency Loss in Machine Learning and High-Dimensional Data}
\label{sec:supp.missing.more.efficiency.loss.in.ml}
There are two reasons why the relative loss of efficiency due to missingness may be greater in typical machine learning settings than in traditional statistical analysis.
First, a statistician might decide to not use a feature that has high missing rates when, based on exploratory analysis, it does not have much predictive power. The typical machine learning workflow would, however, start with all features and rely on automatic feature selection, thereby including the particularly efficiency-damaging features as well. 

Second, for a fixed fraction of missing cells, the larger the number of features, the higher the share of units excluded in complete-case analysis. This is because the more features there are, the higher the probability that an additional missing cell occurs in a row that has no other missing cells and otherwise would not be excluded.
%\highlight{ggf. Grafik}.

\subsection{Aleatoric and Epistemic Uncertainty of Weighting and Imputation Models}
\label{sec:supp.missing.uncertainty.of.weighting.imputation.models}
Weighting and imputation are based on models that themselves are subject to aleatoric and epistemic uncertainty. For weighting, the aleatoric uncertainty is already accounted for by estimating response \textit{propensities}. For imputation, using only a single imputation of a missing feature value via the observed information neglects the aleatoric uncertainty of, e.g., $X_{\text{miss}}|X_{\text{obs}},W$. Hence, in multiple imputation one draws multiple, say $M$, values from the predictive distribution for the missing value, producing $M$ datasets which are analysed separately and whose results are combined according to appropriate rules \citep[p.~8f.]{little.carpenter.lee.2022.soft.intro.to.3.missing.data.method}. 

In deployment with item missingness, consider two units: for the first,  the feature value $x_0$ is actually observed while for the second the same value $x_0$ is the expected value of the imputation. How much uncertainty in the prediction the item missingness adds for the second unit as compared to the first can also be assessed with multiple imputation. We may need more draws $M$ for this purpose than what we are used to from traditional population inference, but generating multiple predictions per unit based on the same prediction model is typically relatively inexpensive computationally.
%much cheaper than having to train multiple prediction models so that higher $M$ should not be prohibitive. 

The epistemic uncertainty in the missing data models is harder to tackle. Resampling-based methods such as bootstrap are sometimes employed for this purpose in traditional statistics \citep[Ch.~3.3.4, 5.3]{little2019statistical}. Embedding the above-mentioned multiple imputation within bootstrapping would account for both aleatoric and epistemic uncertainty in the imputation model. However, because the two epistemic uncertainties -- of the missing data model and of the prediction model -- ``interact'', one would need to train the prediction model anew for each bootstrap iteration of the missing data model. This is typically computationally infeasible unless for the simplest prediction models. 
%In order to still get a (likely underestimated) glimpse, one can treat the prediction model as fixed and ignore the above-mentioned ``interaction''. By embedding the above-mentioned multiple imputation within bootstrapping, 
%of the aleatoric and epistemic uncertainty of each prediction
%to get a (likely underestimated) glimpse of the aleatoric and epistemic uncertainty of each prediction.

Our perception is that the consideration of the two types of uncertainty, aleatoric and epistemic, is not a consistent part of traditional statistical analysis of missing data.

\bibliographystyle{abbrvnat} 
%\bibliography{references}  

\end{localsize}

\end{document}